%% file: main.tex
\documentclass{IEEEtran} 

\usepackage{calc}
\usepackage{indentfirst}
\usepackage{fancyhdr}
\usepackage{lastpage}
\usepackage{ifthen}
\usepackage{lineno}
\usepackage{float}
\usepackage{amsmath}
\usepackage[inline]{enumitem}
\usepackage{mathpazo}
\usepackage{booktabs} 
\usepackage[largestsep]{titlesec}
\usepackage{multirow}
\usepackage{microtype} 
\usepackage{tikz} 
\usepackage{graphicx,epstopdf}

\usepackage{natbib}

\usepackage{url,hyperref}
\usepackage[nopostdot,style=super,nonumberlist]{glossaries}
\usepackage{layouts}
\usepackage{subcaption}
\usepackage[capitalise,noabbrev]{cleveref}
\usetikzlibrary{arrows,positioning}

\makeglossaries
\setacronymstyle{long-short}
\input{glossary}

\def\firstAuthorLast{Nordmoen {et~al.}} 

\def\Address{$^{1}$ Department of Informatics, University of Oslo, Oslo, Norway \\
$^{2}$ RITMO, University of Oslo, Oslo, Norway}

\def\corrAuthor{Jørgen Nordmoen}
\def\corrEmail{jorgehn@ifi.uio.no}

\def\homepage{https://folk.universitetetioslo.no/jorgehn/modular_journal/}
\def\codepage{https://github.com/nordmoen/gym-rem}

\title{MAP-Elites enables Powerful Stepping Stones and Diversity for Modular Robotics}
\author{Jørgen Nordmoen \and Frank Veenstra \and Kai Olav Ellefsen \and Kyrre Glette}
\date{}

\begin{document}
\maketitle
\begin{abstract}
In modular robotics modules can be reconfigured to change the morphology of the robot, making it able to adapt for specific tasks. 
However, optimizing both the body and control of such robots is a difficult challenge due to the intricate relationship between fine-tuning control and morphological changes that can invalidate such optimizations. 
These challenges can trap many optimization algorithms in local optima, halting progress towards better solutions. 
To solve this challenge we compare three different Evolutionary Algorithms on their capacity to optimize high performing and diverse morphologies in modular robotics. We compare two objective-based search algorithms, with and without a diversity promoting objective, with a Quality Diversity algorithm - MAP-Elites. To understand the benefit of diversity we transition the evolved populations into two difficult environments to see if diversity can have an impact on solving more complex environments. In addition, we analyse the genealogical ancestry to shed light on the notion of stepping stones as key to enable high performance.
The results show that MAP-Elites is capable of evolving the highest performing solutions in addition to generating the largest morphological diversity. 
For the transition between environments the results show that MAP-Elites is better at regaining performance by promoting morphological diversity. 
With the analysis of genealogical ancestry we show that MAP-Elites produces more diverse and higher performing stepping stones than the two other objective-based search algorithms.
The experiments transitioning the populations to other more difficult environments show the utility of morphological diversity, 
while the analysis of stepping stones show a strong correlation between diversity of ancestry and maximum performance on the locomotion task.
The paper shows the advantage of promoting diversity for solving a locomotion task in different environments for modular robotics. 
By showing that the quality and diversity of stepping stones in Evolutionary Algorithms is an important factor for overall performance we have opened up a new area of analysis and results.

\end{abstract}

\section{Introduction}

Contemporary research in robotics commonly investigates how to adapt the controllers of robots when exposed to damage or changing environments.
These studies usually consider robots with a fixed morphology. 
In modular robotics, morphological adaptation is achieved through the reconfiguration of modules~\citep{yim2007modular}. With this approach, different morphological configurations can accommodate various tasks and environments~\citep{white2005three}.
However, the possible combinations of modules and control strategies are vast, giving rise to a nontrivial design challenge.

The field of \gls{er} approaches this challenge by applying \glspl{ea} to design and adapt both control and morphology of robots. 
\Glspl{ea} have been successfully applied to modular robot design and control~\citep{hornby2003generative,marbach2004co,faina2013edhmor} although they are prone to premature convergence~\citep{cheney2016difficulty}.
Premature convergence is the phenomenon of having most, or all, solutions in the population converge to local optima, and the prospect of escaping these are difficult without sufficient diversity in the population~\citep{hornby2006alps}.  
This challenge is compounded when evolving \emph{modular} robots due to the connection between controller optimization and morphology. In one relevant study, \citet{faina2013edhmor} observe a high degree of deceptiveness in the search landscape when evolving modular robots, leading standard \glspl{ea} to underperform.
The available variation operators, such as adding a module, may easily invalidate the current control strategy~\citep{cheney2016difficulty}.


Overcoming challenges in modular robotics require optimization algorithms that are able to evolve high performing solutions while retaining morphological diversity to avoid premature convergence in the resulting deceptive fitness landscapes. 
While little research exists so far on this challenge in the context of modular robotics, proposed approaches include a custom constructive approach~\citep{faina2013edhmor}, morphological protection mechanisms~\citep{cheney2018scalable}, or introducing a controller learning phase for new morphologies~\citep{jelisavcic-frontiers2019}.
These techniques essentially allow the morphology and control to change on different time scales.

Recent advances in \gls{er} are based on promoting phenotypic diversity in the search process. 
One simple but powerful way to achieve this is by making phenotypic difference in the current population an additional objective to maximize~\citep{mouret2009overcoming}. 
This multi-objective approach utilizes traditional \glspl{moea} where one objective is the traditional performance value and another objective is added to represent the diversity of solutions~\citep{mouret2011novelty}.
\Gls{qd} is an emerging paradigm within the field of \glspl{ea}~\citep{pugh2016quality}.
This class of algorithms go beyond the singular focus on maximizing one or more objectives and instead actively construct a repertoire of phenotypically unique and high-performing solutions~\citep{cully2017quality}. 

\Gls{qd} algorithms differentiate solutions based on phenotypic properties, also called behavioral descriptors, which dictate the inclusion into an archive.
The most popular variants of \gls{qd} algorithms are \gls{nslc}~\citep{lehman2011evolving}, which uses an unstructured archive, and \gls{map-elites}~\citep{mouret2015illuminating}, which uses a structured archive: an N-dimensional grid spanning the behavioral descriptor space.
Focusing on novelty instead of fitness alone has shown to find solutions in deceptive fitness landscapes~\citep{lehman2008exploiting}, and \Gls{qd} algorithms have successfully been applied to evolve diversity of virtual creatures~\citep{lehman2011evolving}.
%
In~\citep{miras2018effects}, a morphological novelty measure was included in a composite fitness function when evolving modular robots.
However, the study focused on the diversity of the resulting morphologies, and locomotion performance was negatively affected compared to pure performance-based fitness function.
Consequently, the application of a complete \gls{qd} approach to tackle the above-mentioned challenges of evolving morphology and control for modular robots warrants exploration. 

One potential reason for the efficacy of \gls{qd} algorithms is the notion that \gls{qd} algorithms are better at promoting and exploiting stepping stones~\citep{mouret2015illuminating}. 
In the context of \glspl{ea}, a stepping stone is a solution that other better solutions build upon. 
In that way, a stepping stone does not need to have any other quality apart from being in the genealogical ancestry of the concluding solution. 
In \citep{mouret2015illuminating} the authors propose that \gls{map-elites} is better at finding high performing solutions because the search algorithm is better at promoting diverse stepping stones. 
Another comparison of \gls{qd} algorithms and objective-based search for the generation of stepping stones can be found in~\citep{gaier2019quality}. 
Here the authors argue that due to the ability of \gls{map-elites} to promote poor, but `novel' solutions, that can later be built upon to become good solutions, it is able to overcome the premature convergence experienced with the objective-based approach. 
This suggests that analysing the potential of \gls{qd} algorithms for generating stepping stones could be a way to increase our knowledge about this class of search algorithms and help explain the difference between \gls{qd} algorithms and objective-based search methods.

Building on our initial study in \citep{nordmoen2020quality}, this paper compares three \glspl{ea} on their ability to evolve high performing and morphologically diverse modular robots. 
We utilize two objective-based search algorithms, one without a diversity objective and one with a diversity objective, and the \gls{qd} algorithm \gls{map-elites}~\citep{mouret2015illuminating} to illuminate the difference between these two paradigms as applied to the modular robotics domain. 
Our goal is to understand how the morphological difference evolved with these three search algorithms affect the task, when the environment changes and different morphological needs arise.
Furthermore, to understand the algorithmic differences, we present an analysis of the genealogical ancestry of the evolved populations to shed light on the hypothesis that \gls{qd} algorithms perform better due to a difference in how stepping stones are generated and utilized. 
To achieve the stated goals we created a new modular robotics framework \gls{rem} which is used to simulate and evolve the modular robots for this paper.

The contributions of our paper are three-fold:
First we demonstrate that \gls{map-elites} is well suited for the difficult task of evolving both morphology and control in modular robotics.
By extending our previous results, we show that differing the selection pressure can have a large impact on maximum fitness obtained for this \gls{qd} algorithm. 
We expand on the performance results by transitioning the populations of the three search algorithms between two different environments showing that as environmental complexity grows, the necessity for morphological diversity increases. 
Secondly, we present a way of analysing the genealogical ancestry of all three algorithmical approaches to better understand how stepping stones can lead to different results. 
By looking at the statistical properties of the ancestry we gain the ability to generalize over all experimental runs which increases the confidence in our results. 
Finally, in addition to the two previous contributions, we release a new framework for evolving modular robotics, the \gls{rem} framework, which leverages OpenAI Gym and PyBullet to achieve fast and easy to extend simulations, opening up modular robotics to a wider machine learning audience.

\subsection{Related Work}
\subsubsection{Modular Robotics}
Evolving body and control for artificial creatures have a long history in the field of Artificial Life~\citep{sims1994evolving}. 
Modular robotics is distinguished from these virtual creatures by comprising the morphology of re-usable homogeneous or heterogeneous building blocks, called modules~\citep{Stoy2010, moubarak2012modular}.
This is in contrast to virtual creatures where individual body parts can evolve to have any shape and size.
By using these building blocks, modular robotics provide a way to effectively transition from simulation to reality as modules can be fabricated individually and then combined based on designs optimized in simulation~\citep{stoy2006deformatron}.
By designing modules in such a way as to make them easy to build in the real world, modular robotics offers a great deal of freedom for optimization in simulation since simulated robots can easily be put together from pre-built parts and transitioned to the real world for performance verification~\citep{moreno2017emerge}. 
Through reusing modules and recent advances for potentially auto-assembling modular robots \citep{Brodbeck2015,Moreno2018,Hale2019}, this approach can become more feasible since the robot does not need to be constructed from the ground up. 

One challenge in modular robotics is the interconnected relationship between control and morphology~\citep{lipson2000automatic, cheney2016difficulty}.
To overcome this challenge many different approaches such as generative encodings~\citep{hornby2003generative,veenstra2017evolution} and different control architectures~\citep{marbach2005online,haasdijk2010hyperneat} have been applied.

\subsubsection{\texorpdfstring{\acrlong*{qd}}{Quality Diversity}}
\Gls{qd} algorithms emerged from the realization that optimization through promoting phenotypic diversity can yield high performing solutions and, more importantly, can be better suited to exploring the whole problem space~\citep{lehman2008exploiting}.
Through actively searching for phenotypic diversity, \gls{qd} algorithms traverse the search space without constraining the search to only finding better-fit solutions~\citep{pugh2016quality}. 
This separates \gls{qd} algorithms from traditional \glspl{moea} since Pareto dominated solutions can be kept as long as their phenotypic expression is sufficiently different from other solutions in the population~\citep{mouret2015illuminating}.
An interesting property of \gls{qd} algorithms is the capability to produce a repertoire of different solutions for the same problem~\citep{cully2017quality}. 
The repertoire can be exploited, either at design time~\citep{gaier2017data} or during operation~\citep{cully2015robots}, to select different solutions depending on the circumstances of the situation.

Although \gls{qd} algorithms have been applied to the evolution of artificial creatures~\citep{lehman2011evolving} and morphological descriptors have been
used to evolve robots~\citep{samuelsen2014some, samuelsen2015real} few examples exist applying the \gls{qd} paradigm to modular robotics.
A related area of inspiration is voxel-based soft robotics~\citep{hiller2012dynamic}. Several works have explored soft robot design with \gls{qd} algorithms such as \citep{methenitis2015novelty} which first applied novelty search, \citep{gravina2018fusing} which combines novelty- and suprise search and \citep{gravina2019blending} which compares different forms of diversity with \gls{map-elites} in the soft robotics domain.

\section{Materials and Methods}
\subsection{\texorpdfstring{\acrlong*{rem}}{Robotics, Evolution and Modularity} Framework}
For our experiments we created a new simulation framework based on PyBullet~\citep{coumans2016pybullet} and OpenAI Gym~\citep{brockman2016openai}, called \acrfull{rem}~\footnote{Source available at: \url{\codepage}}. 
PyBullet is the Python interface to the Bullet~\citep{coumans2015bullet} physics simulator and OpenAI Gym is a framework to standardize simulations, initially within the reinforcement learning domain, that prescribes a few necessary functions that together create and run a simulation. 
OpenAI Gym makes it easy to reproduce setups from different experiments through exposing multiple environments through a common interface. 
Through using OpenAI gym we thereby make our framework more accessible for people, especially those already familiar with OpenAI gym.
By building the framework on PyBullet the orchestration code can be programmed in Python while Bullet itself is written and optimized in \textit{`C'}. 

\begin{figure*}
    \centering
    \begin{subfigure}[b]{0.45\textwidth}
        \includegraphics[width=\textwidth]{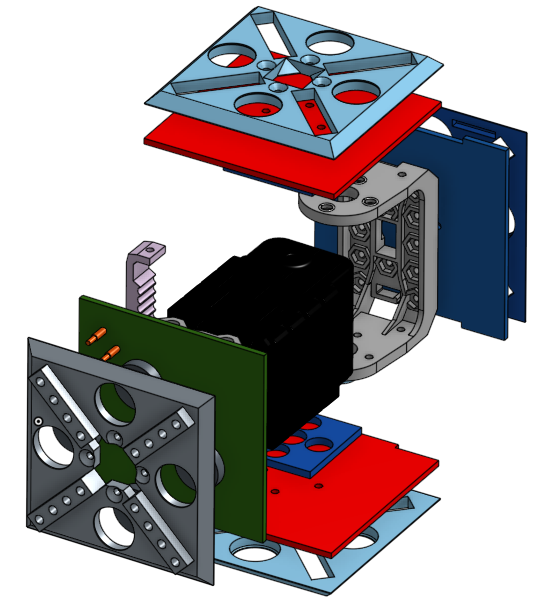}
    \end{subfigure}
    \begin{subfigure}[b]{0.45\textwidth}
    \includegraphics[width=\textwidth]{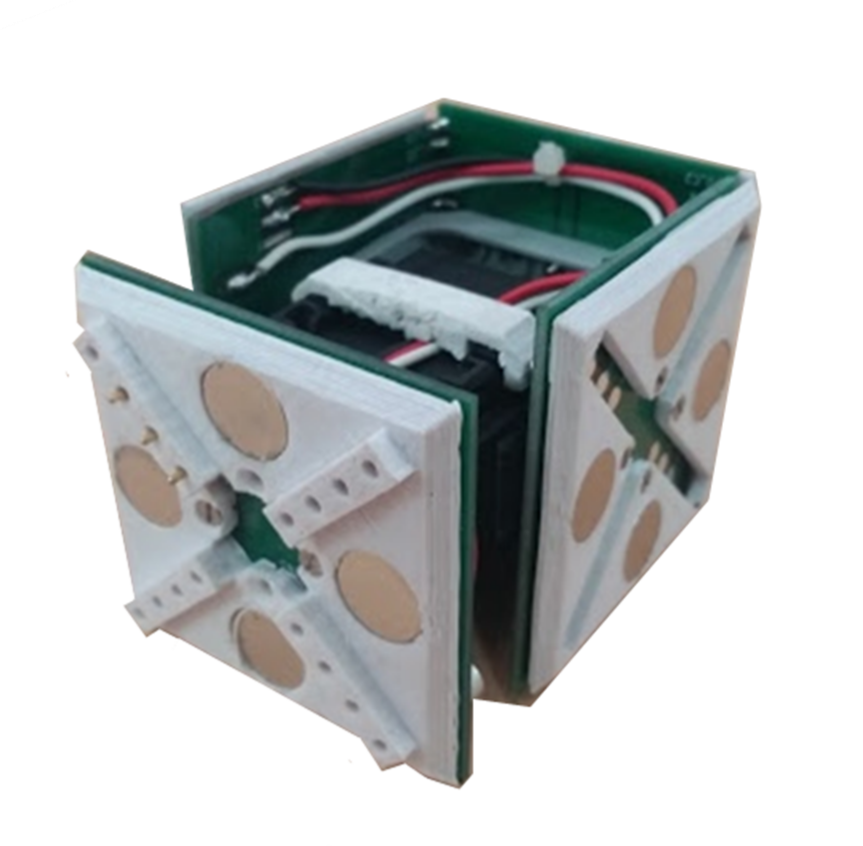}
    \end{subfigure}
    \caption{The real world module which \gls{rem} utilizes as joint module. On the \textbf{left} is an exploded view of the CAD model and on the \textbf{right} is the assembled real module \citep{moreno2017emerge}}. 
    \label{fig:real_module}
\end{figure*}

The modules supported by \gls{rem} are based on the EMeRGE~\citep{moreno2017emerge} modules and have real-world properties for size, weight and joint forces. 
At the time of publication, the \gls{rem} framework supports two different module types, one movable joint module which is based on Dynamixel AX-18\textemdash shown in \cref{fig:real_module}, and a non-movable module with the same dimensions as the joint module sans joint. 
The connection between modules are based on magnets which makes the real-world modules easy to assemble and disassemble. 
Unfortunately Bullet does not support connections that can break at a certain force threshold and so this is not supported yet. 
Simulation is default performed at $240Hz$, when graphical interface is not enabled, to give a high degree of accuracy for simulation.

To support different genome encodings the \gls{rem} framework utilizes a tree-based acyclic graph representation for phenotypes which describes the morphology to instantiate. 
This allows for both direct and indirect genome encodings as long as they unpack into the graph expected by \gls{rem}.

\subsection{Encoding and Control System}
\begin{figure*}
    \centering
    \begin{subfigure}[b]{0.45\textwidth}
    \include{figures/encoding}
    \end{subfigure}
    \begin{subfigure}[b]{0.45\textwidth}
    \includegraphics[width=\textwidth]{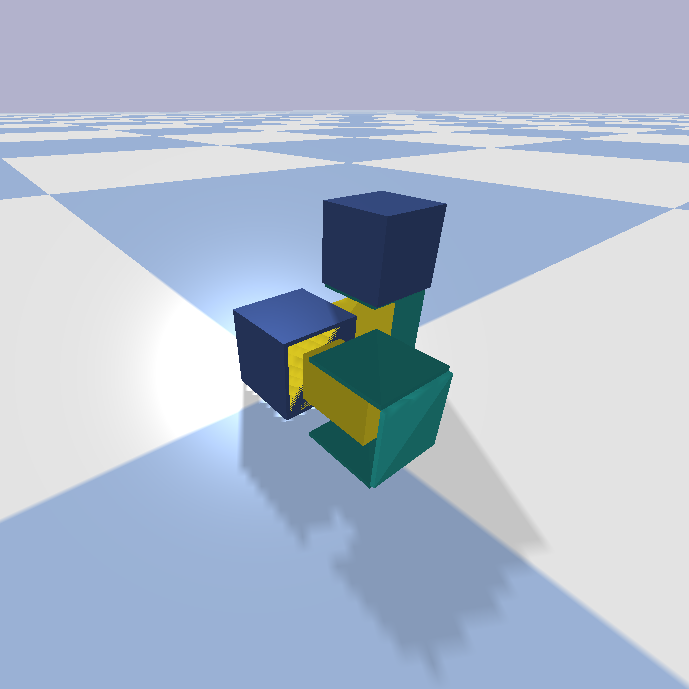}
    \end{subfigure}
    \caption{On the \textbf{left} the tree representation of the encoding is shown. The nodes in the graph represent modules and edges represent connections between modules. Each module has a certain number of possible connections where the triple $(X, Y, Z)$ denote where the connection is, relative to the parent, in 3D. Note that modules have additional properties, such as \textit{rotation}, which is not shown in this illustration. On the \textbf{right} the simulated modular robot corresponding to the encoding is shown in the \gls{rem} framework.}
    \label{fig:encoding}
\end{figure*}

The morphological encoding employed in the experiments is a tree-based direct encoding similar to~\citep{faina2013edhmor}, illustrated in \cref{fig:encoding}.
The encoding allows for any directed acyclic graph of modules to be represented, where each node in the graph represents a module and each edge is a connection between two modules. 
The encoding corresponds one-to-one with the phenotype encoding in the \gls{rem} framework.
For the experiments carried out in this article two different modules were utilized, one non-movable rectangular module supporting $5$ child modules and one servo module capable of moving one side back-and-forth and supporting $3$ child modules~\citep{moreno2017emerge}. 
Each morphology starts with a single rectangular module as its root. To randomly initialize the morphology, a random size is selected between $1$ and $\eta$ (see~\cref{tab:morph_parameters}).
Modules are then added to the tree at random locations until the size of the morphology equals the selected size.

The morphological encoding supports mutation- and crossover-operators. When mutating the morphology three possibilities exist:
\begin{enumerate*}
    \item \textit{Add a random module.} The tree is traversed and each available
            connection point is added as a possibility. A
                connection point is randomly selected along with a randomly
                selected module type before being inserted into the tree.
    \item \textit{Remove a module.} The tree is traversed adding all modules,
            except the root, into a list of candidates to remove. A module is
                randomly selected from the candidates before being removed along
                with any existing children.
    \item \textit{Mutate a module.} The two modules in use both support rotation
            around its connection axis and mutation will randomly select a new
                orientation in $90^o$ increments.
\end{enumerate*}
Note that \textit{only} one of the three possibilities can happen per morphological mutation.

For crossover, a branch exchange is implemented. For both parent morphologies the tree is traversed adding all modules, except the root, to a list of candidates. 
A random candidate is selected from both morphologies before being exchanged. The candidate module, including its children, from the first morphology, is inserted into the place of the candidate from the second morphology and vice versa.

Lastly, the morphology is limited to a maximum size, $\eta$ (see~\cref{tab:morph_parameters}), and a maximum depth $\delta$, so that additional modules are not realized in the simulator. This limit ensures that morphologies do not grow unbounded and are feasible to simulate.

\begin{table}
\caption{Morphological parameters for the evolved modular robots.}
\label{tab:morph_parameters}
\centering
\begin{tabular}{ccc}
\hline
\textbf{Parameter} & \textbf{Description} & \textbf{Value}\\
\hline
$\eta$ & Maximum module count & 20\\
$\delta$ & Maximum module depth from root & 4\\
\hline
\end{tabular}
\end{table}

The control system of the joint modules are based on a decentralized wave pattern generator~\citep{veenstra2017evolution}. 
Each joint module is initialized with a controller that is updated on each simulation tick to output desired angle of the joint, $\theta_i$, according to the following equation

\begin{equation}
    \theta_i = \alpha_i * \sin\left(\omega_i t + \phi\right) + o_i
\end{equation}

where $\alpha_i$ is the amplitude, $t$ is the time since the controller was initialized, $\omega_i$ is the frequency, $\phi_i$ is the phase offset and $o_i$ is the amplitude offset. The output of the controller, i.e. the maximum and minimum values of $\theta_i$, is limited so it does not exceed the ability of the real world module. The parameters and their allowable ranges are defined in \cref{tab:ctrl_parameters}.

\begin{table}
\caption{Parameters for the decentralized wave pattern controllers. The value ranges are based on the servo used in the real world modules.}
\label{tab:ctrl_parameters}
\centering
\begin{tabular}{ccc}
\hline
\textbf{Parameter} & \textbf{Description} & \textbf{Range}\\
\hline
$\theta$ & Set-point angle & $[-1.57, 1.57]$\\
$\alpha$ & Amplitude & $[-1.57, 1.57]$\\
$\omega$ & Frequency & $[0.2, 2]$\\
$\phi$ & Phase offset & $[-2\pi, 2\pi]$\\
$o$ & Amplitude offset & $[-1.57, 1.57]$\\
\hline
\end{tabular}
\end{table}

The controllers are mutated using Gaussian noise, $\mathcal{N}(p, \sigma)$, where $p$ is the individual parameter of the controller and $\sigma$ is the magnitude of the noise. The magnitude, $\sigma$, is scaled for each parameter so that a global mutation rate can be used for the controller, the scaling is defined by the range of each parameter detailed in \cref{tab:ctrl_parameters}. To avoid mutating values outside their defined bounds we utilize the \textit{bounce-back} restriction function~\citep{nordmoen2020restricting}.

\subsection{Evolutionary Algorithms}
To better understand how \gls{qd} algorithms are able to evolve both high performance and diverse solutions we will compare three different \glspl{ea} on the task of evolving both control and morphology in modular robotics. 
The comparison will initially utilize a flat terrain environment before experimenting in more complex simulated environments. 
The fitness objective of the \glspl{ea} is the straight-line distance traversed, between the initial starting point and final position of the robot, during the evaluation. 
For the diversity preserving \glspl{ea} morphological properties will be used to distinguish solutions. 
Selecting which morphological properties to utilize is a challenging problem~\citep{miras2018search}. 
In this paper we define the number of non-movable and the number of movable joint modules as morphological properties that can be used as our diversity metric. 
By selecting these simple features we focus on the search algorithms and not the morphological features~\citep{samuelsen2014some}.

\begin{figure*}
    \begin{subfigure}[b]{\textwidth}
        \centering
        \includegraphics{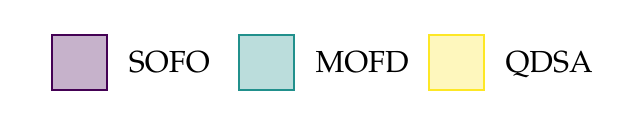}
    \end{subfigure}
    \begin{subfigure}[b]{\textwidth}
        \centering
        \includegraphics{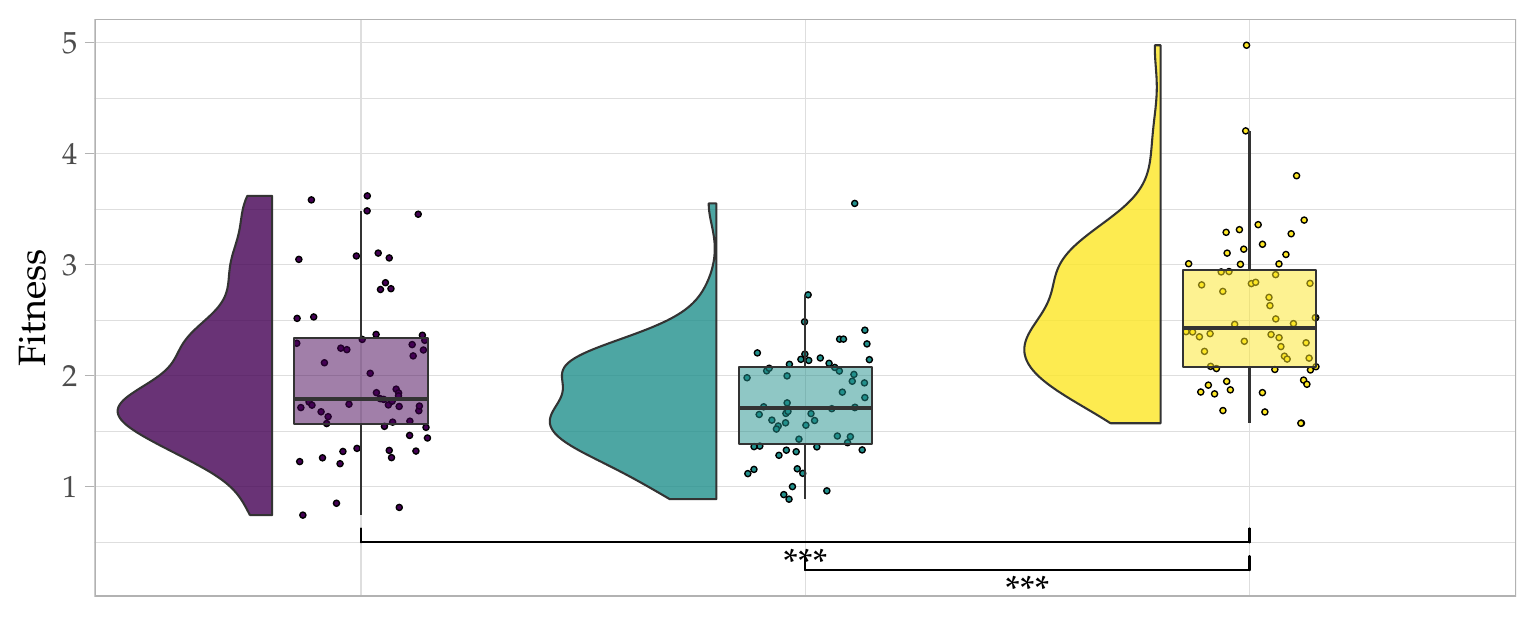}
    \end{subfigure}
    \caption{Summary of parameter optimization. All 60 runs for each search algorithm are shown together with the result of a Mann-Whitney U test which show statistical significant differences between \gls{map} and the two other search algorithms.}
    \label{fig:param_sweep}
\end{figure*}

To ensure a balanced comparison, the mutation parameters of each algorithm were optimized in advance in a parameter sweep (\cref{tab:meta_parameters}). 
Each parameter was tested twice for each \gls{ea} and $100\ 000$ evaluations were done for each set of combined parameters. 
The simulation time for each evaluation was limited to 20 seconds in these initial runs. 
In total $180$ runs were conducted to ascertain the best parameters for each search algorithm.
Based on these results, a linear model was constructed to predict fitness based on the interaction of the two parameters tested. 
The best parameter combination of each algorithm was chosen to be used in the remaining experiments of this paper. 
A summary of all the runs is shown in \cref{fig:param_sweep}, which shows that \gls{map} is on average slightly better than the two other search algorithms\textemdash regardless of parameter combination.

\begin{table}
\caption{Values tested during meta-optimization of experiment parameters.}
\label{tab:meta_parameters}
\centering
\begin{tabular}{cc}
\hline
\textbf{Parameter} & \textbf{Values}\\
\hline
Probability of morphological mutation & $[0.005, 0.01, 0.05, 0.1, 0.2, 0.4]$\\
Controller magnitude ($\sigma$) & $[0.005, 0.01, 0.05, 0.1, 0.2]$\\
\hline
\end{tabular}
\end{table}

The first \gls{ea} is a single objective, $(\lambda, \mu)$ generational replacement strategy, based on~\citep{eiben2003introduction}. 
The algorithm optimizes for fitness alone and is used as a baseline to compare the two other diversity preserving algorithms. 
The algorithm utilizes tournament selection between two solutions, based on fitness, for selection and incorporates elitism, preserving $10$ of the best solutions from the previous generation. 
For the rest of this article we will refer to this algorithm as \gls{single}. 
Experiment parameters for this algorithm can be found in \cref{tab:exp_parameters}.

The first of the diversity preserving algorithms is the \gls{moea}, \gls{nsga}~\citep{deb2002fast}. 
This \gls{ea} also represents the objective optimization perspective, however, since \gls{nsga} is capable of optimizing multiple objectives the algorithm can be used to optimize for diversity~\citep{lehman2011evolving}. 
The diversity metrics used are based on morphological descriptors of the evolved robots and comprise the number of non-movable modules and the number of movable modules. 
In order for \gls{nsga} to optimize for diversity, the average difference between morphologies is used as an objective~\citep{lehman2011evolving}, according to the following equations:

\begin{equation}\label{eq:diversity}
    diversity(x) = \frac{1}{|P_n|} \sum_{y \in P_n} distance(x, y)
\end{equation}

\begin{equation}\label{eq:distance}
    distance(x, y) = 1.0 - e^{|(m_x, j_x) - (m_y, j_y)|}
\end{equation}

where $P_n$ is the population, $x$ and $y$ are solutions in the population, $m_i$ is the number of non-movable modules and $j_i$ is the number of movable joint modules. 
Note that the output of both equations is given in $\mathbb{R}^2$, which gives three objectives for \gls{nsga} to optimize. 
Note also that the distance equation is altered compared to previous work~\citep{samuelsen2014some, lehman2011evolving} to avoid convergence at the morphological extremities. 
By changing the distance function (\cref{eq:distance}) to using the natural exponential function all changes in morphology are weighted equally, which prevents large changes in morphology from dominating the diversity calculation during optimization. 
In effect, adding one or ten modules is weighted as equally diverse. 
For the rest of this article we will refer to this algorithm as \gls{multi}. 
Experiment parameters for the algorithm can be found in \cref{tab:exp_parameters}.

The last \gls{ea} used represents the \gls{qd} paradigm and is the \gls{map-elites} algorithm~\citep{mouret2015illuminating}. 
Central to the \gls{map-elites} algorithm is the archive, or repertoire, which is utilized to store and select solutions. 
The archive is structured with cells of equal sizes that represent a specific combination of feature descriptors~\citep{cully2017quality}. As with \gls{multi}, we utilize morphological properties as feature descriptors.
In contrast to \glspl{moea}, \gls{map-elites} does not utilize multiple objectives, however, diversity is promoted through the archive by allowing multiple solutions to be differentiated by their feature descriptors. 
For the experiments carried out in this article, the archive consists of two dimensions where one axis represents the number of non-movable modules and the other axis represents the number of movable joint modules. 
The dimensions are scaled to the maximum size of a morphology, as described in \cref{tab:morph_parameters}, and the cell at the origin represents the root module. 
For selection we utilize tournament selection based on the \textit{curiosity} of solutions in the repertoire~\citep{cully2017quality}. 
Curiosity is implemented by adding $1.0$ to the curiosity score of a parent when a child is inserted into the repertoire and subtract $0.5$ when a child fails to be inserted into the repertoire, which corresponds to the values suggested in~\citep{cully2017quality}.
For consistency, we will refer to this search algorithm as \gls{map}. 
Parameters used for experiments for this algorithm can be found in \cref{tab:exp_parameters}. 

\begin{table*}
\caption{Experiment parameters for the search algorithms.}
\label{tab:exp_parameters}
\centering
\begin{tabular}{ccc}
\hline
\textbf{Parameter} & \textbf{Applied to} & \textbf{Value}\\
\hline
Evaluation time & \multirow{7}{*}{All} & 20 seconds\\
Warm-up before start & & 2 seconds\\
Repetitions & & 30 \\
Number of evaluations & & 100 000\\
Batch size & & 200\\
\hline
Probability of crossover & & 0.2\\
Probability of controller mutation & & 1.0 \\
\hline
\multirow{3}{*}{Initial population size} & \acrshort{single} & \multirow{2}{*}{200}\\
& \acrshort{multi} &\\
& \acrshort{map} & 1000\\
\hline
\multirow{3}{*}{Selection} & \acrshort{single} & \multirow{2}{*}{Tournament on objective(s)}\\
& \acrshort{multi} &\\
& \acrshort{map} & Tournament on \textit{curiosity}\\
\hline
\multirow{3}{*}{Probability of morphological mutation} & \acrshort{single} & \multirow{2}{*}{0.2}\\
& \acrshort{multi} &\\
& \acrshort{map} & 0.4\\
\hline
\multirow{3}{*}{Controller mutation magnitude ($\sigma$)} & \acrshort{single} & \multirow{2}{*}{0.01}\\
& \acrshort{multi} &\\
& \acrshort{map} & 0.005\\
\hline
\end{tabular}
\end{table*}

\subsection{Objectives}
The environment where the robots have to move in shape the search space of our experiments. 
With a flat terrain, an evolutionary run might lead to a smooth progression due to the absence of obstacles/ deceptive traps.
To see whether all the approaches perform the same when changing the environment, we use three different environments: A flat terrain, a raised platform with a single wall, and a circular terrain where circular walls ripple outwards (\autoref{fig:envs}).

\section{Results}
\subsection{Performance and Diversity}
\begin{figure*}
    \centering
    \begin{subfigure}[b]{\textwidth}
        \centering
        \includegraphics{cache/legend.pdf}
    \end{subfigure}
    \begin{subfigure}[b]{0.45\textwidth}
        \includegraphics{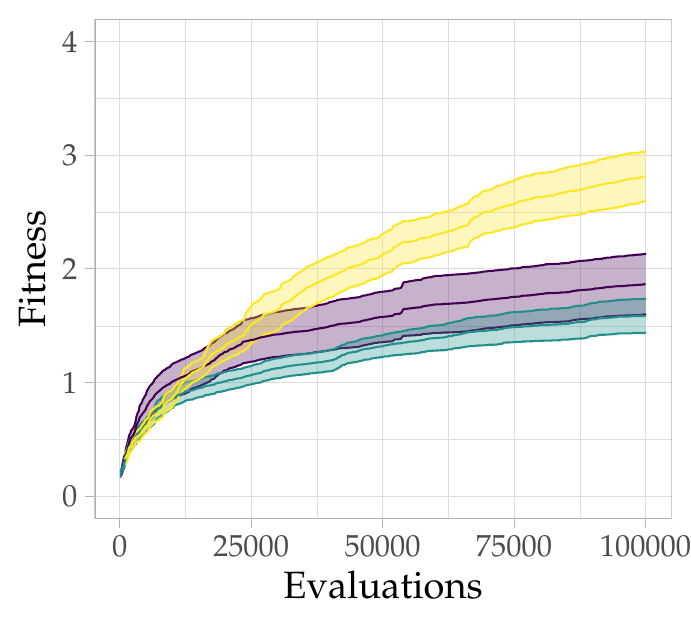}
    \end{subfigure}
    \begin{subfigure}[b]{0.45\textwidth}
        \includegraphics{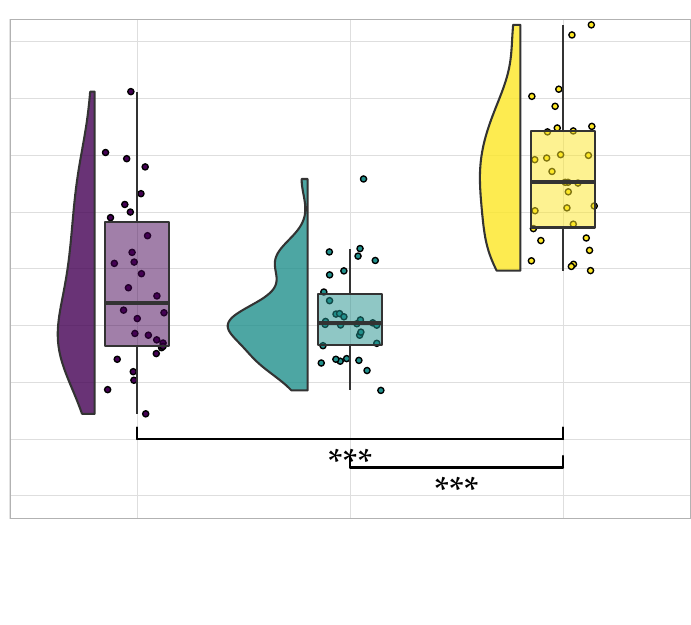}
    \end{subfigure}
    \caption{Fitness of the single best solution found in the population. On the \textbf{left} the mean is shown together with a $95\%$ confidence interval over generational time. On the \textbf{right}, the best fitness after the last evaluation for all repetitions is shown. Statistically significant differences are marked on the right using a Mann-Whitney U test with Holm correction.}
    \label{fig:fitness}
\end{figure*}

To begin analysing the performance of the three search algorithms, we will start by looking at the best fitness obtained by any single solution in the population. 
The best fitness is plotted in \cref{fig:fitness}, where on the left the fitness is shown over the number of performed evaluations, and on the right the single best individual found after the last evaluation is shown. 
A Mann-Whitney U test~\citep{mann1947} between the three distributions in the right plot of \cref{fig:fitness}, corrected for multiple comparison through Holm correction~\citep{holm1979simple}, shows that there is a significant difference between \gls{map} and the two other search algorithms. 
For locomotion, \gls{map} is able to find the best performing solution of the three search algorithms.

\begin{figure*}
    \centering
    \includegraphics{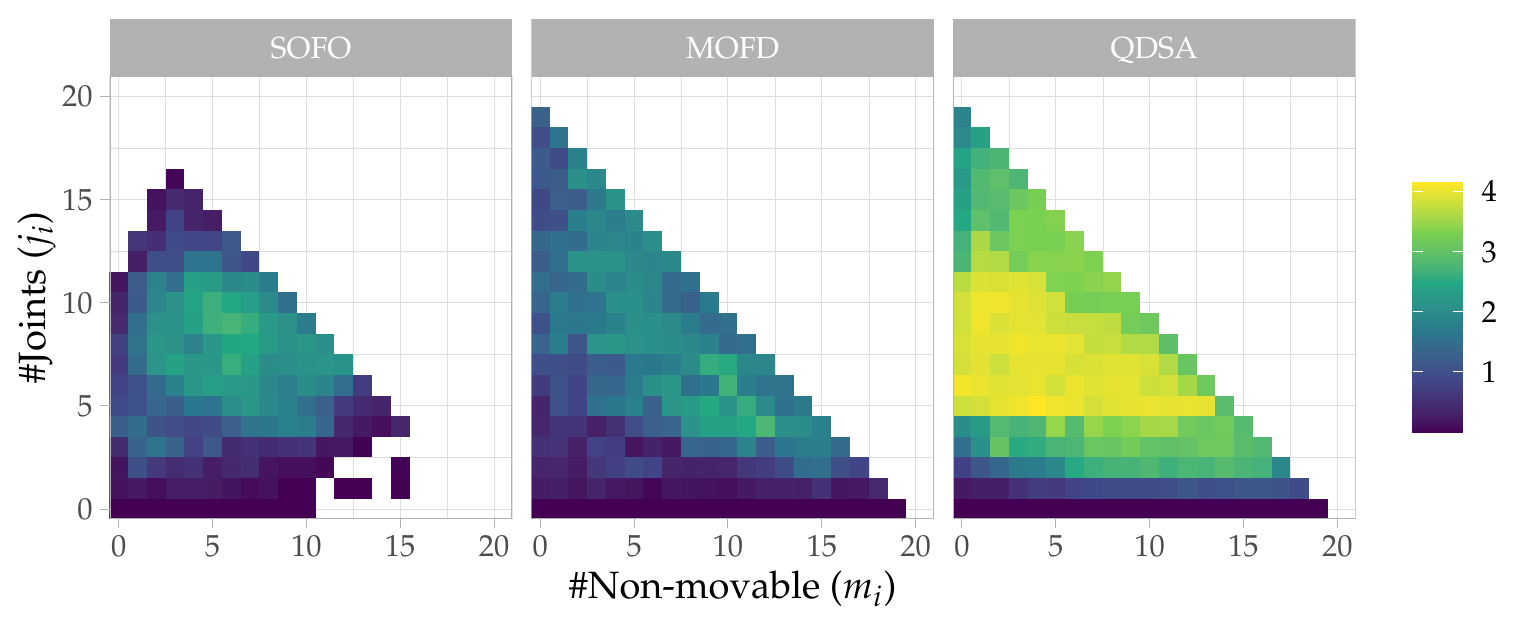}
    \caption{Maximum fitness of morphological niches for each search algorithm over $30$ runs as projected into a repertoire with the morphological descriptors as axes. The color represents fitness with a range shown in the colorbar on the right. Note that all morphologies contain a non-movable root module.}
    \label{fig:population_max}
\end{figure*}

\begin{figure*}
    \centering
    \includegraphics{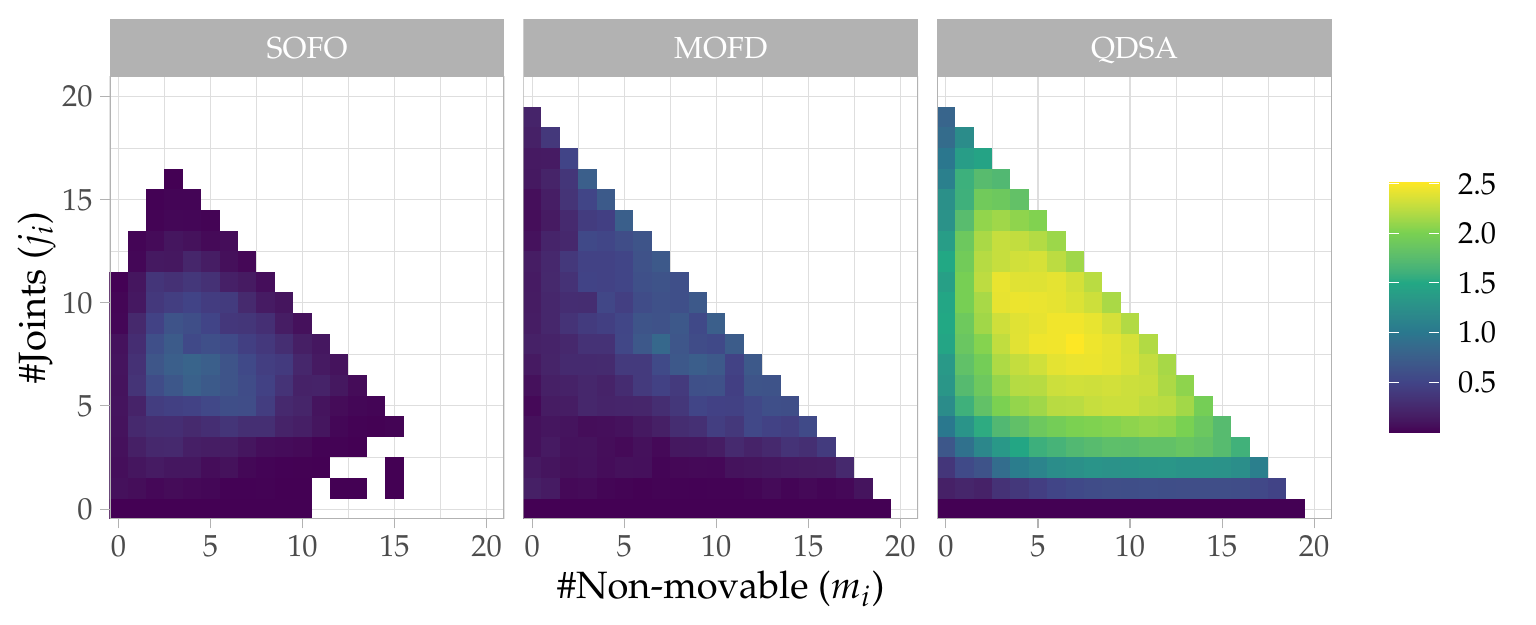}
    \caption{Average fitness of morphological niches for each search algorithm over $30$ runs as projected into a repertoire with the morphological descriptors as axes. The color represents fitness with a range shown in the colorbar on the right. Note that all morphologies contain a non-movable root module.}
    \label{fig:population_avg}
\end{figure*}

Since it is not only the fitness, or quality, of the solutions that we are interested in it is informative to project the population of solutions into a repertoire using the morphological descriptors as axes. This projection gives an overview of the population as a whole and it is possible to visualize where in the morphological space the best solutions are found. The projection also enables us to visualize the quality-diversity trade-off, which shows that not all solutions, or morphologies, can obtain the same fitness. In \cref{fig:population_max} the maximum fitness for each morphological niche is shown, while \cref{fig:population_avg} shows the average fitness. The two figures illustrates the difference in diversity between the different search algorithms and shows how consistent the algorithms are at discovering solutions.

\begin{figure*}
    \centering
    \includegraphics{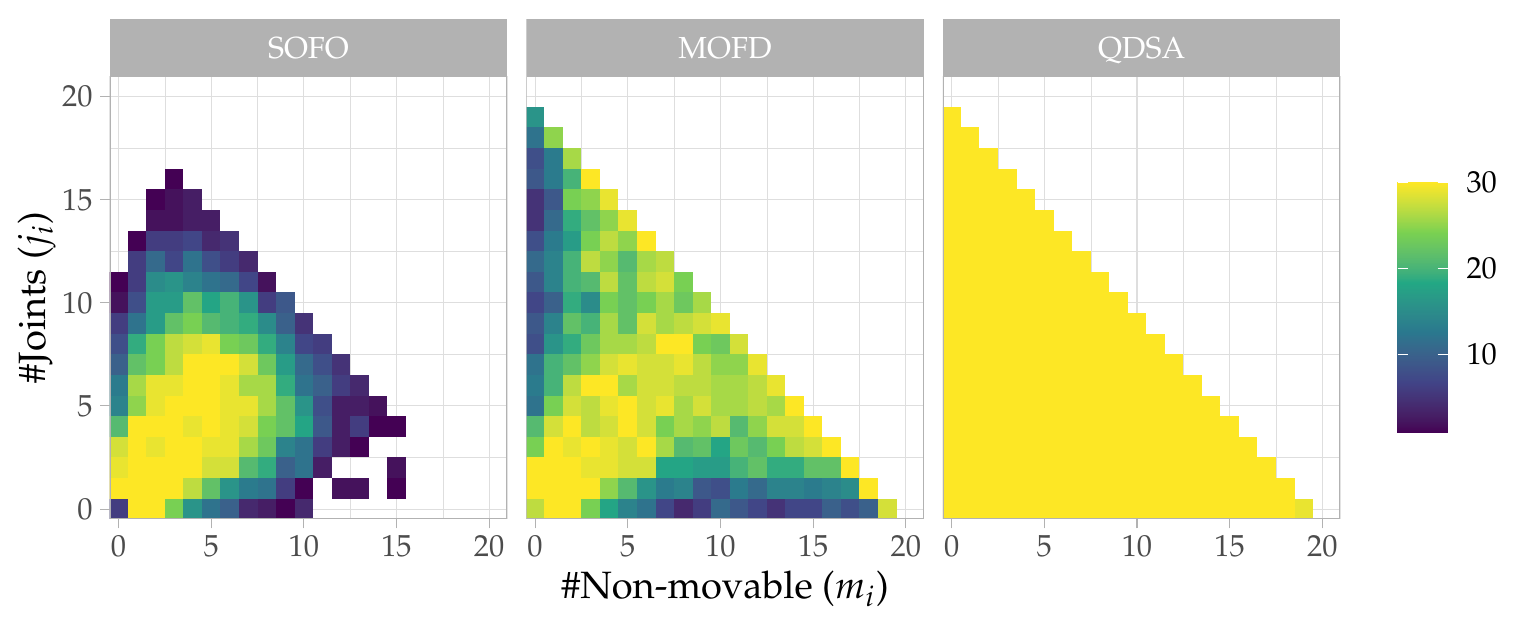}
    \caption{Number of repetitions that found a solution for the given morphological descriptor. Color represents the number of runs which found any solution for the given morphological description. Note that all morphologies contain a non-movable root module.}
    \label{fig:population_num}
\end{figure*}

Although \cref{fig:population_avg} shows diversity of the search algorithms, through the average fitness of morphologies, it is not able to show how proficient each algorithm is at finding diverse solutions, low performance may just indicate that the morphological niche cannot perform better. To alleviate this, \cref{fig:population_num} shows the number of experiments which found a solution for each morphological niche. From the figure it can be seen that \gls{map} and \gls{multi} are more consistent in finding diverse solutions while the single objective \gls{single} is centered around a smaller cluster of morphologies. Although \gls{multi} is able to more find more diverse solutions than \gls{single} it can be seen that only \gls{map} consistently finds solutions for all niches.

\begin{figure*}
    \centering
    \begin{subfigure}[b]{\textwidth}
        \centering
        \includegraphics{cache/legend.pdf}
    \end{subfigure}
    \begin{subfigure}[b]{0.45\textwidth}
        \includegraphics{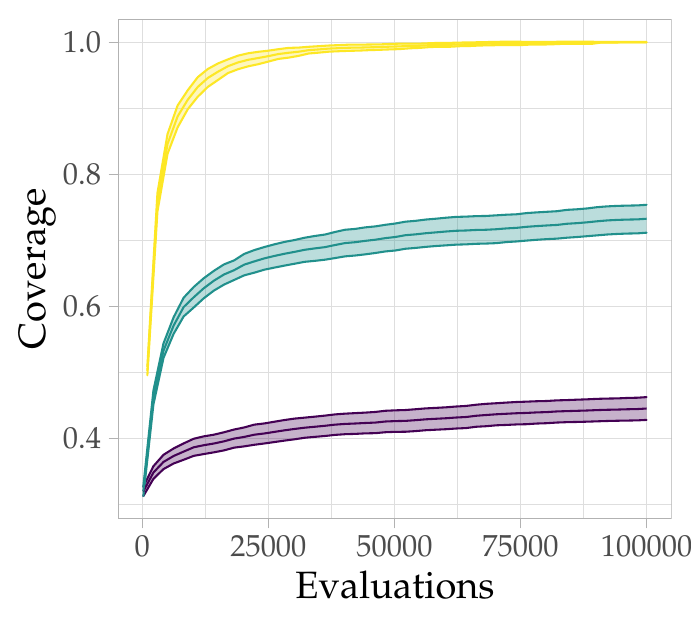}
    \end{subfigure}
    \begin{subfigure}[b]{0.45\textwidth}
        \includegraphics{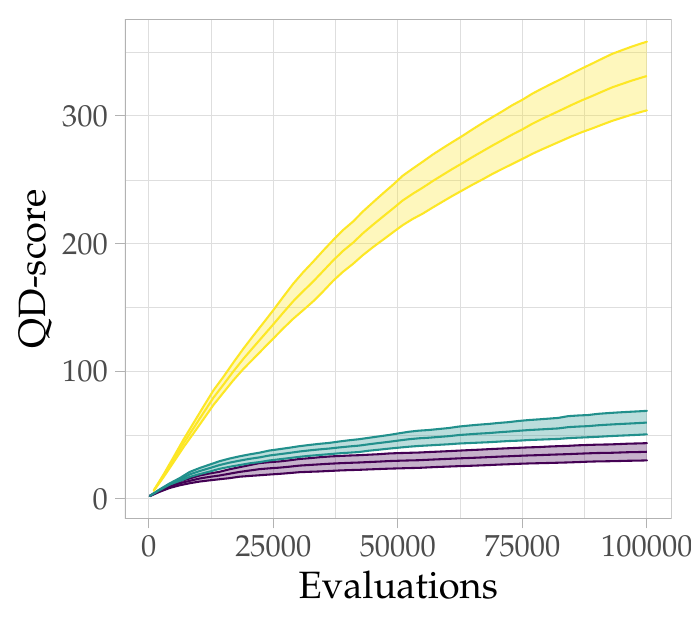}
    \end{subfigure}
    \caption{QD-Metrics. On the \textbf{left} the normalized coverage~\citep{mouret2015illuminating}, number of discovered morphological niches, is shown. Coverage has been normalized to the maximum number of niches found in any run for all three search algorithms. On the \textbf{right} the \gls{qd}-score~\citep{pugh2016quality}, summation of fitness for all solutions in the population, is shown.}
    \label{fig:qd_metrics}
\end{figure*}

To summarize the projections in \cref{fig:population_avg,fig:population_num} we can utilize metrics suggested by~\citet{mouret2015illuminating} and~\citet{pugh2016quality}. 
\Cref{fig:qd_metrics} shows the coverage and \gls{qd}-score of the three search algorithms. 
Coverage counts the number of unique niches found in the population and is normalized to the maximum coverage found in any run of all algorithms. Coverage can be viewed as a summation of the data shown in \cref{fig:population_num} and shows the evolved diversity of the search algorithms. 
\gls{qd}-score is the sum of fitness of each solution in the population and is a good summation of the quality and diversity trade-off. 
\Gls{qd}-score gives a more balanced view than either \textit{precision} or \textit{reliability} since both of these metrics decrease as new low fitness solutions are added, due to the lower average performance and thus disadvantaging search algorithms that generate diversity. 
A Mann-Whitney U test shows that the differences between all three search algorithms, after the last evaluations, for both plots in \cref{fig:qd_metrics} are significant. 
The two graphs in \cref{fig:qd_metrics} show the complexity of comparing algorithms on the trade-off between quality and diversity, even though \gls{multi} has a much higher coverage compared to \gls{single} the difference in \gls{qd}-score is much lower due to \gls{single} having on average high fitness in the niches it fills out.

\begin{figure*}
    \centering
    \begin{subfigure}[b]{0.3\textwidth}
    \includegraphics[trim={10cm 10cm 10cm 10cm}, clip, width=\textwidth]{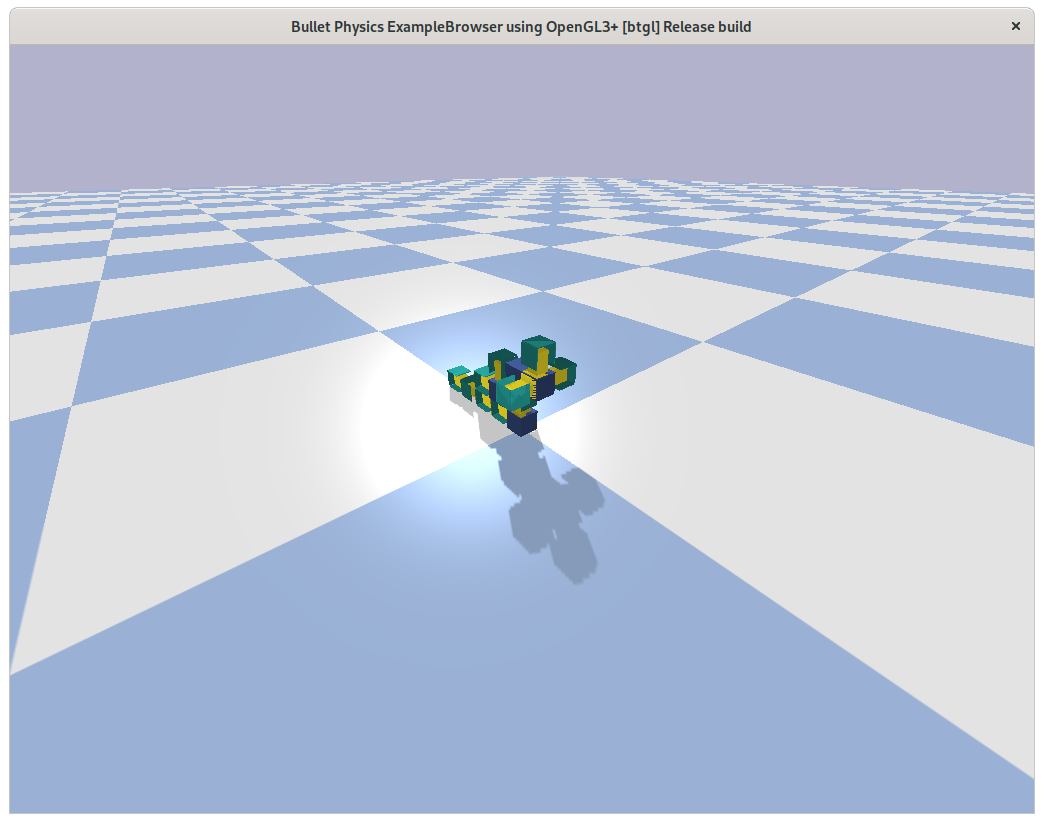}
    \end{subfigure}
    \begin{subfigure}[b]{0.3\textwidth}
    \includegraphics[trim={10cm 10cm 10cm 10cm}, clip, width=\textwidth]{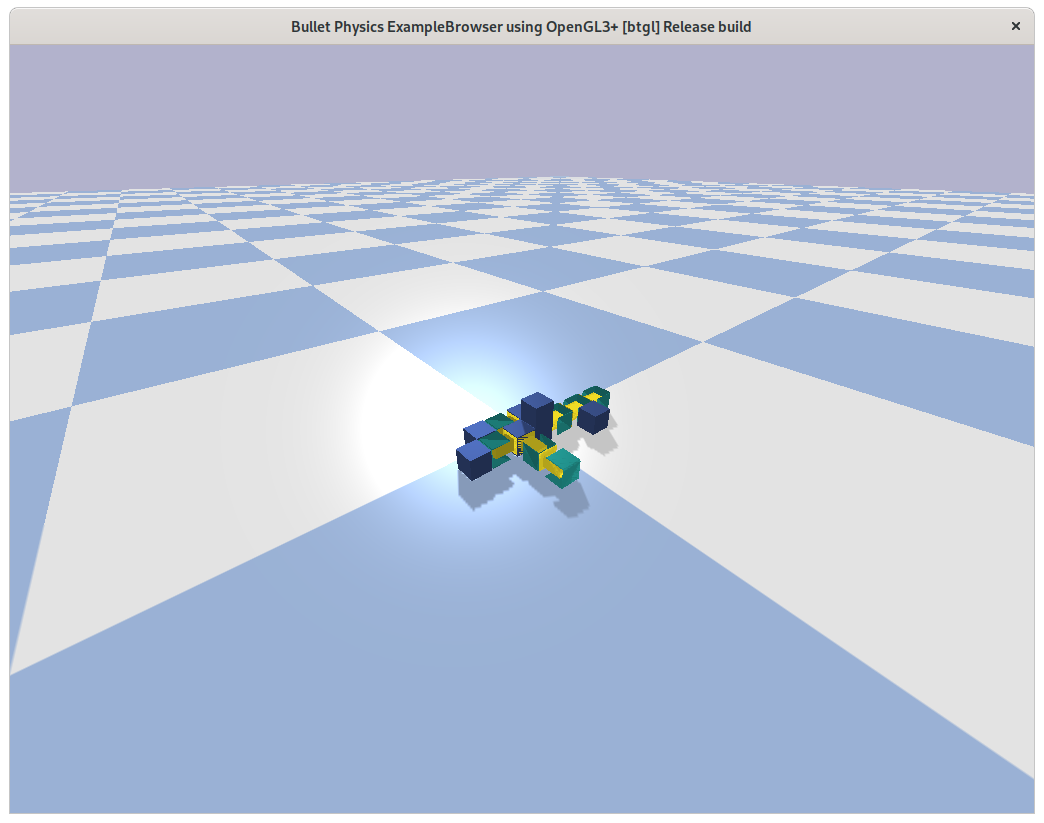}
    \end{subfigure}
    \begin{subfigure}[b]{0.3\textwidth}
    \includegraphics[trim={10cm 10cm 10cm 10cm}, clip, width=\textwidth]{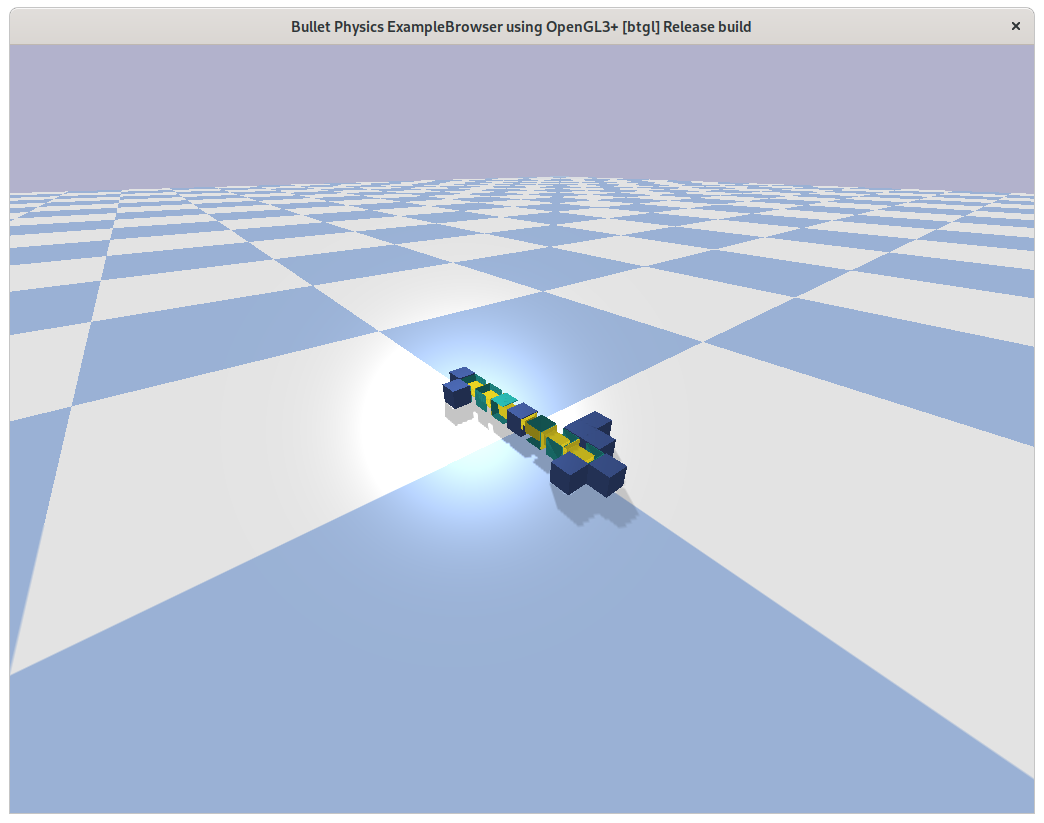}
    \end{subfigure}
    \begin{subfigure}[b]{0.3\textwidth}
    \includegraphics[trim={10cm 10cm 10cm 10cm}, clip, width=\textwidth]{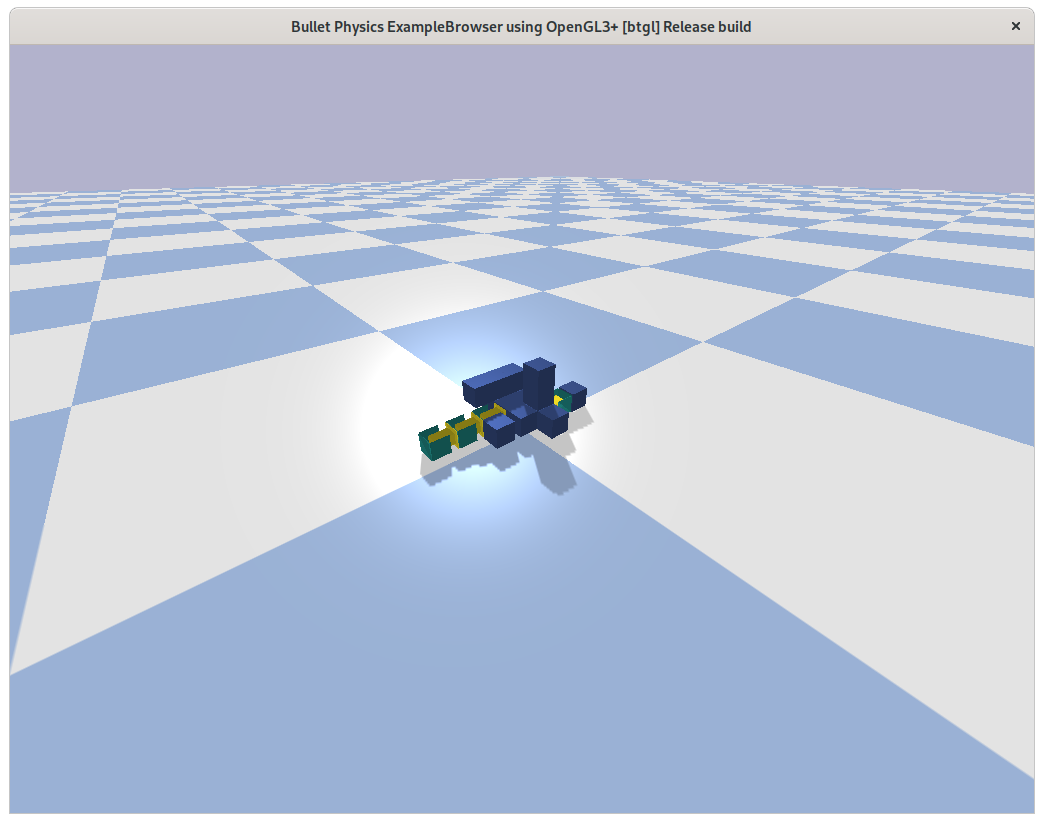}
    \end{subfigure}
    \begin{subfigure}[b]{0.3\textwidth}
    \includegraphics[trim={10cm 10cm 10cm 10cm}, clip, width=\textwidth]{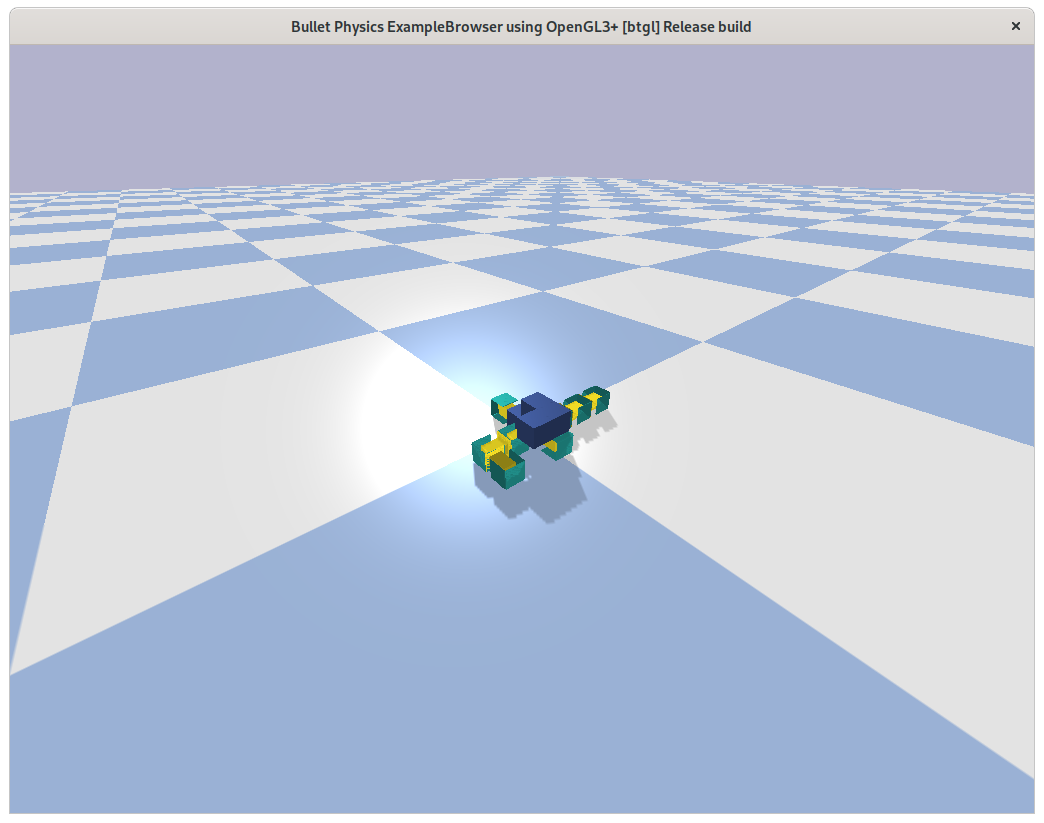}
    \end{subfigure}
    \begin{subfigure}[b]{0.3\textwidth}
    \includegraphics[trim={10cm 10cm 10cm 10cm}, clip, width=\textwidth]{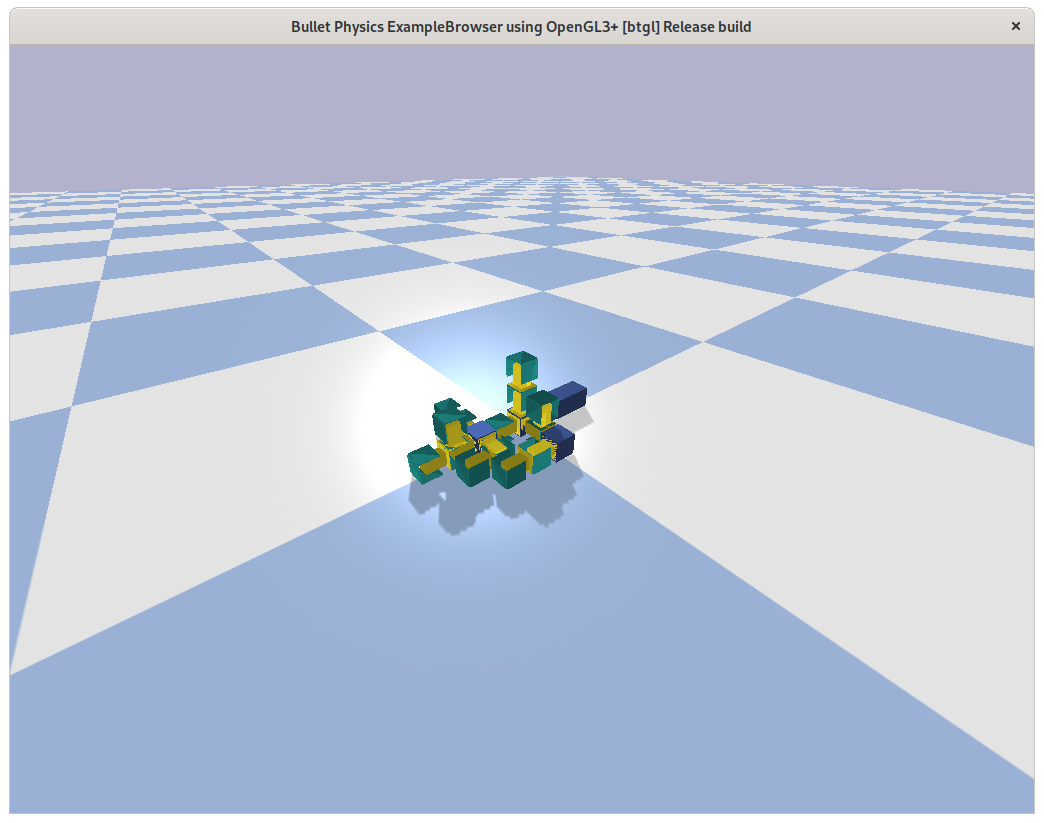}
    \end{subfigure}
    \begin{subfigure}[b]{0.3\textwidth}
    \includegraphics[trim={10cm 10cm 10cm 10cm}, clip, width=\textwidth]{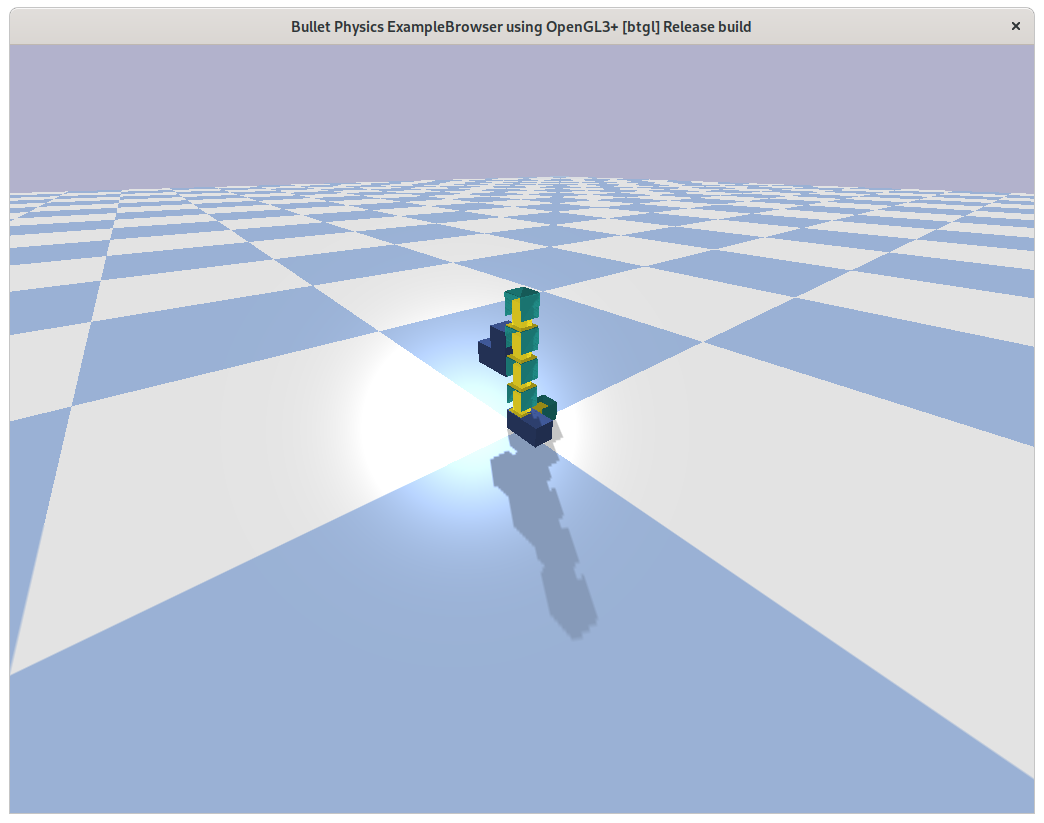}
    \end{subfigure}
    \begin{subfigure}[b]{0.3\textwidth}
    \includegraphics[trim={10cm 10cm 10cm 10cm}, clip, width=\textwidth]{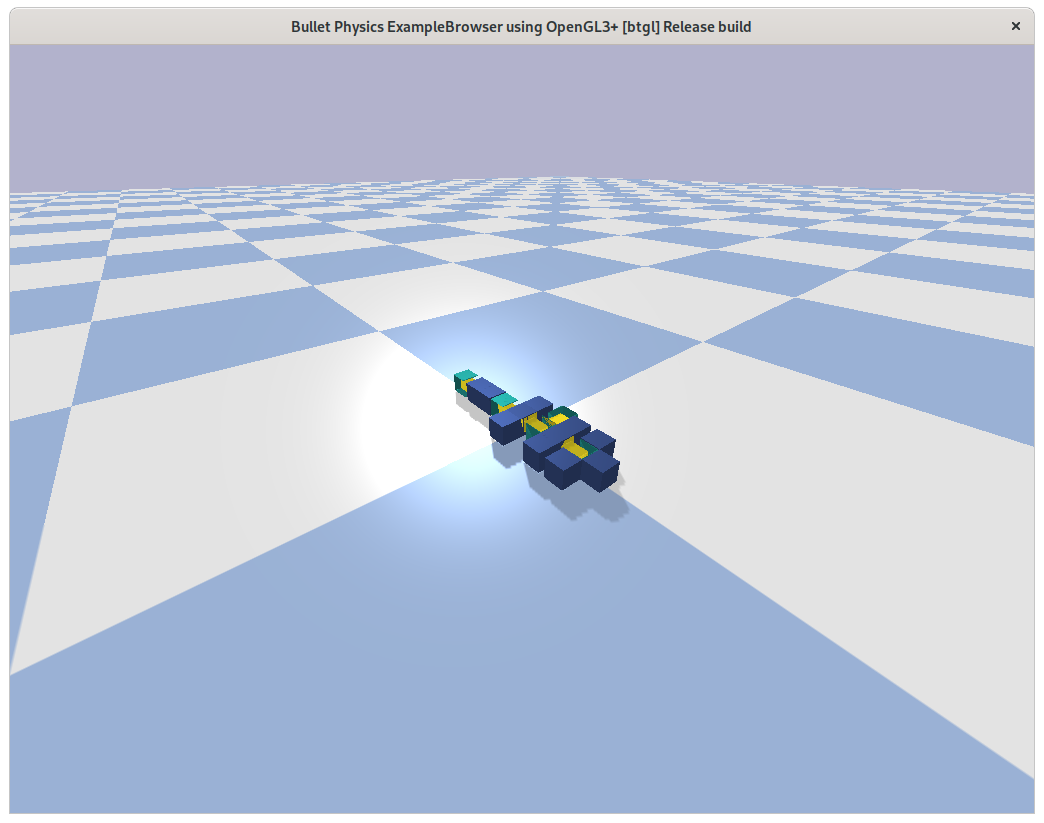}
    \end{subfigure}
    \begin{subfigure}[b]{0.3\textwidth}
    \includegraphics[trim={10cm 10cm 10cm 10cm}, clip, width=\textwidth]{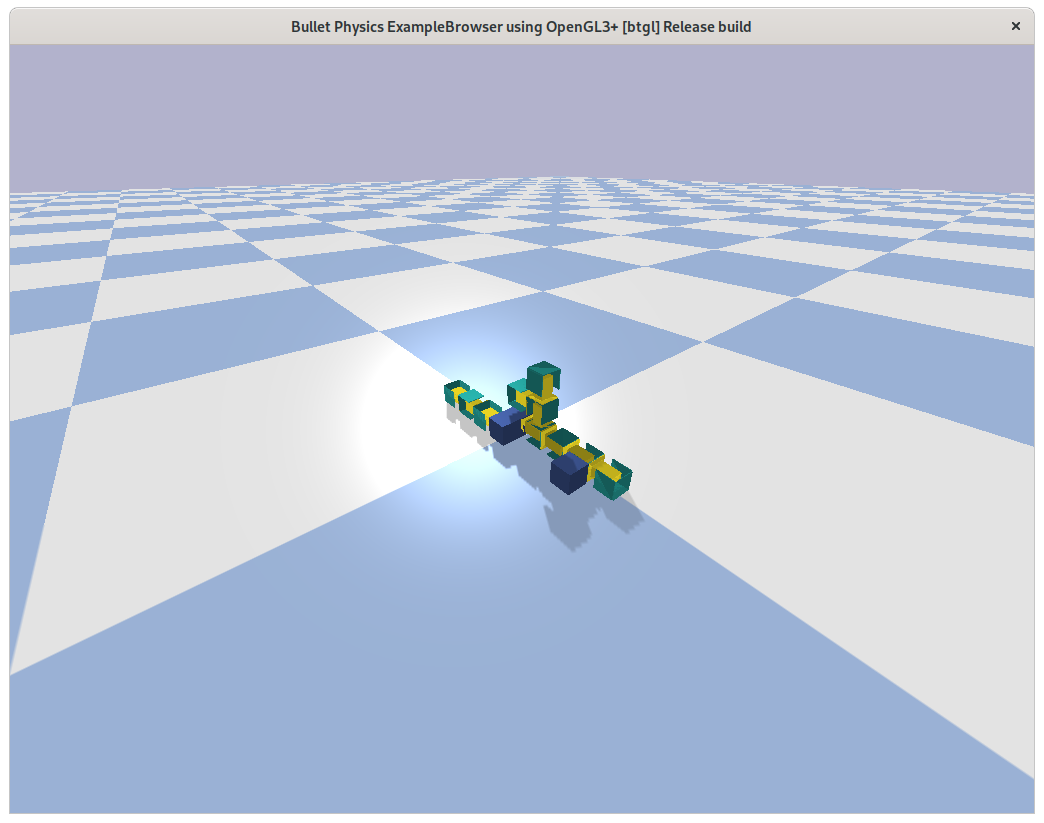}
    \end{subfigure}
    \caption{The best solution found in the three best runs of evolution for each search algorithm. The \textbf{top row} shows \gls{single}, the \textbf{middle row} shows \gls{multi} and the \textbf{bottom row} shows \gls{map}. Videos of these morphologies can be found on the supplementary material page.}
    \label{fig:robots}
\end{figure*}

To get an impression of the evolved morphologies, we selected the three best runs from each search algorithm and extracted the single best solution. The solutions are shown in \cref{fig:robots}. Although the morphologies look quite similar, just rotated about the $Z$-axis differently, they are quite diverse once one counts the differences in modules and types.

\subsection{Transitioning to New Environments}
\begin{figure*}
    \centering
    \begin{subfigure}[b]{0.45\textwidth}
    \includegraphics[trim={0 0 0 10cm}, clip, width=\textwidth]{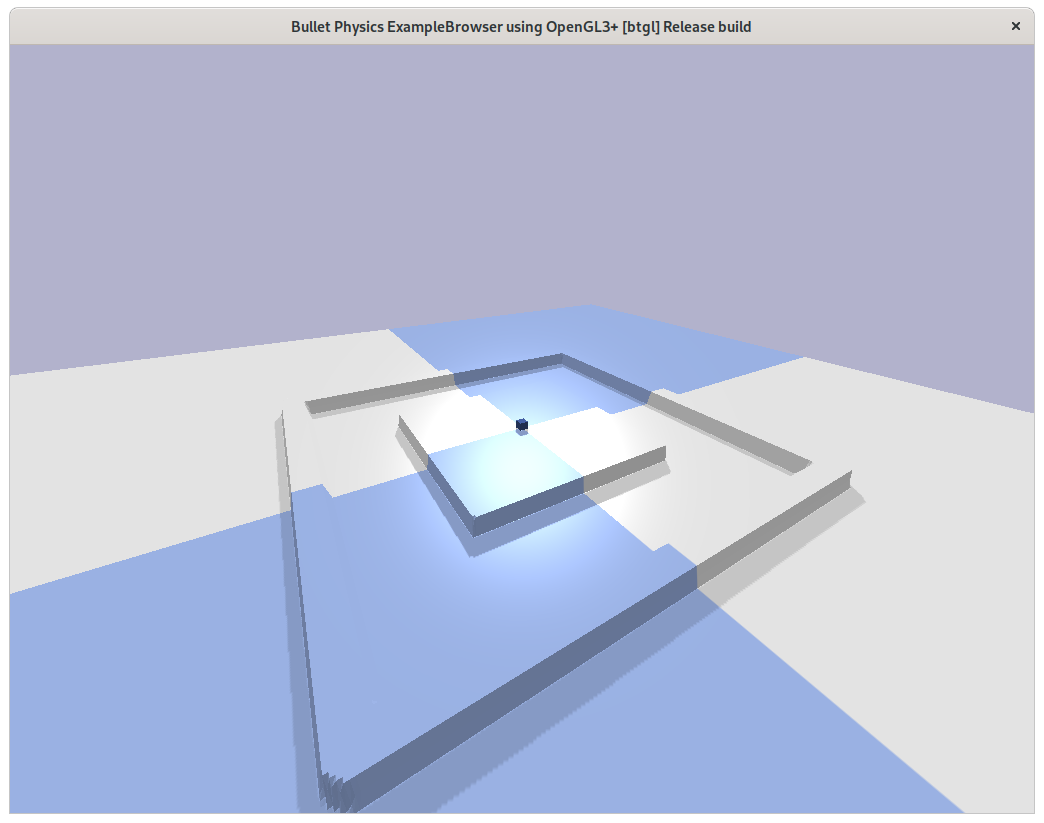}
    \end{subfigure}
    \begin{subfigure}[b]{0.45\textwidth}
    \includegraphics[trim={0 0 0 10cm}, clip, width=\textwidth]{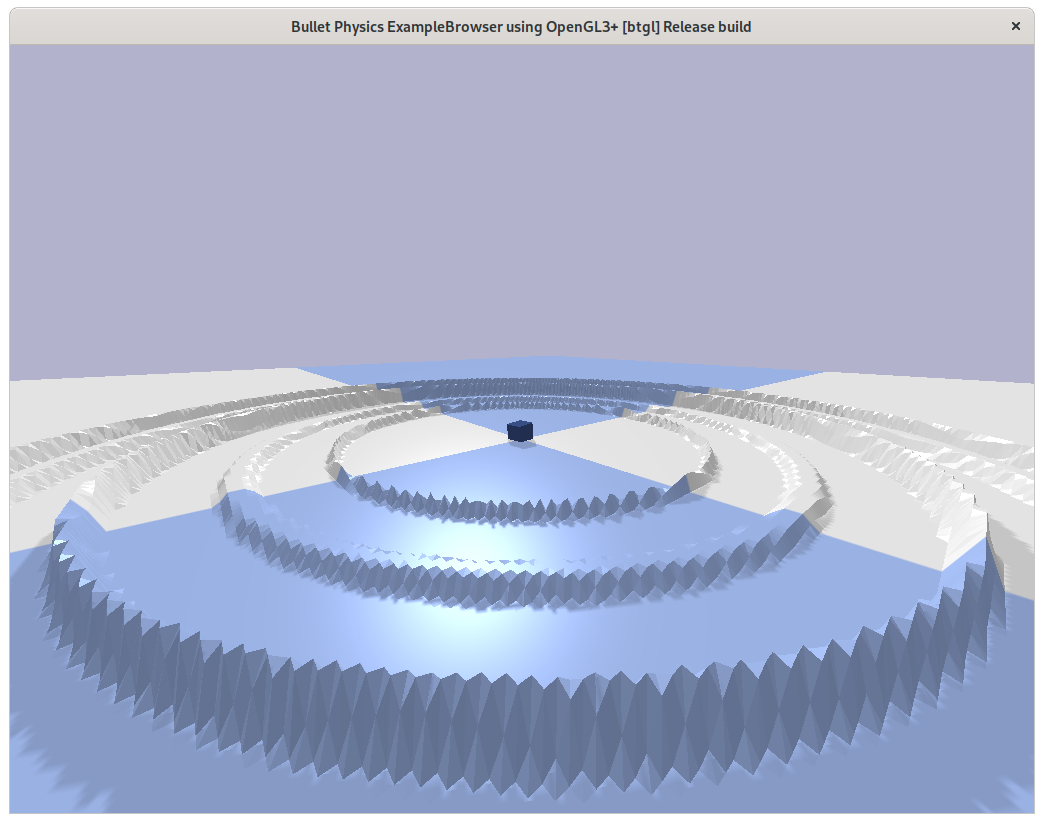}
    \end{subfigure}
    \caption{The more complex environments used to test transferred solutions. On the \textbf{left} the robots start on a raised platform until a ditch is created with a single wide wall. On the \textbf{right} the robots start on a flat terrain and several circular thin walls ripple outwards becoming taller and taller.}
    \label{fig:envs}
\end{figure*}

\begin{figure*}
    \centering
    \begin{subfigure}[b]{\textwidth}
        \centering
        \includegraphics{cache/legend.pdf}
    \end{subfigure}
    \begin{subfigure}[b]{\textwidth}
        \centering
        \includegraphics{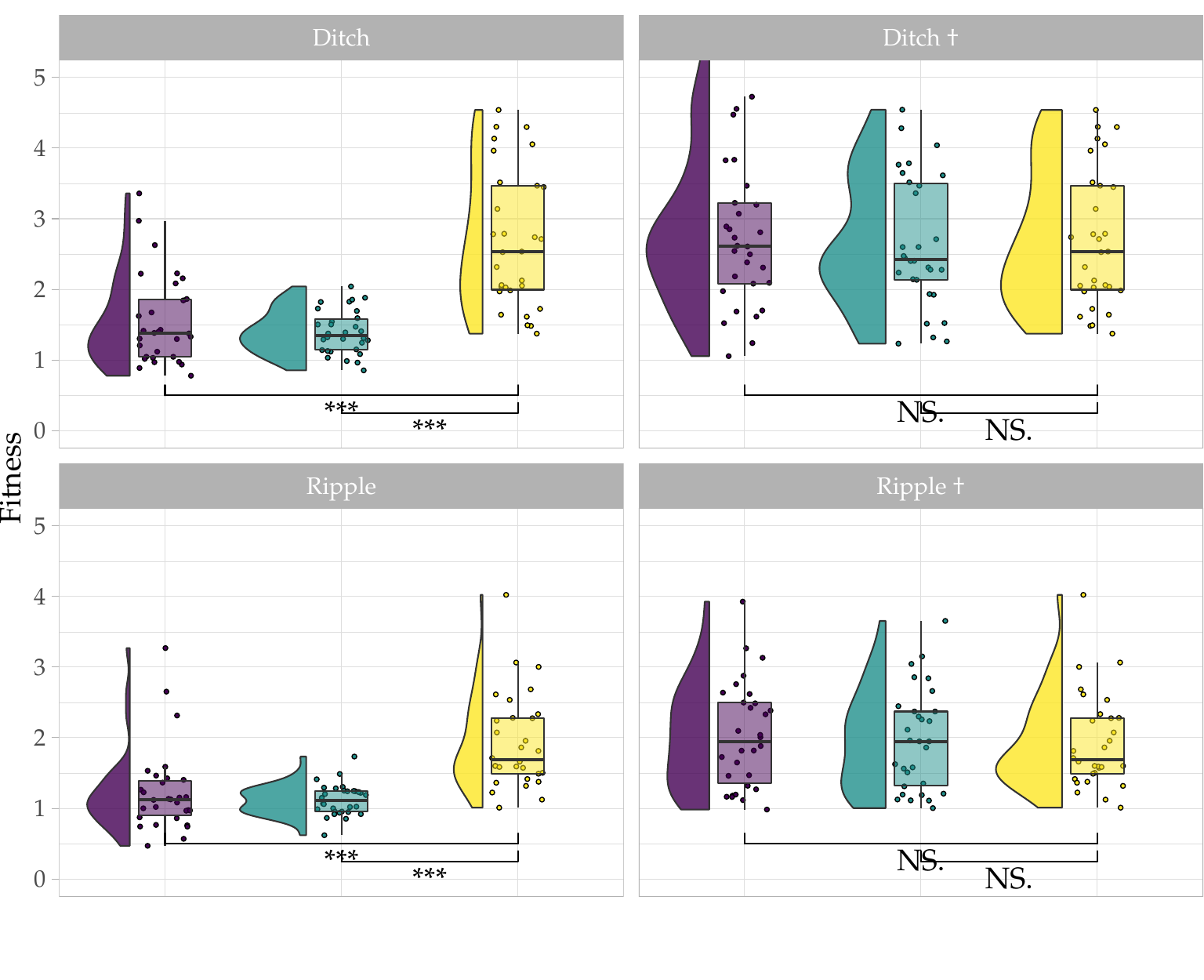}
    \end{subfigure}
    \caption{Fitness results after last evaluation for transitioning the population evolved in the default flat environment into two new environments. In the new environment evolution is continued from the initial seed population for $50 000$ evaluations. Names with $\dagger$ signify that both \gls{single} and \gls{multi} was initialized with the result of \gls{map} from the default flat environment.}
    \label{fig:env_last}
\end{figure*}

To better understand the value of diversity in our modular robotics scenario we created two new environments, with different obstacle profiles as shown in \cref{fig:envs}, to see if the difference in evolved morphologies would lead to differing result in more challenging environments. The hypothesis being that a more diverse population should transition better into a different environment since an already discovered morphology could potentially lead to good performance in the new environment. Said in another way, convergence to a few good solutions in one environment could lead to slow evolution in another environment if none of the converged solutions are able to solve the new environment. We tested this hypothesis by transitioning the final population in \cref{fig:fitness} into two different environments and `continue' evolution from the population evolved for the default environment. The results are shown in the left column of \cref{fig:env_last}. From the left column it can be seen that \gls{map} is able to obtain the best fitness in both new environments. The difference between \gls{single} and \gls{multi} is not significantly different, however, it is interesting to note that they seem to have changed relative position compared to \cref{fig:fitness}, a change that could indicate that the diversity of \gls{multi} is aiding in transitioning into a new environment. In addition to testing the result of each search algorithm we also tested if the population evolved in the default environment with \gls{map} could aid the other two algorithms. The population of \gls{map} was transitioned from the default environment into the two new environments, but instead of using as mentioned in \citep{cully2017quality} the two other search algorithms were utilized to continue evolution. The results of continuing from the population of \gls{map} in the two new environments are shown in the right column of \cref{fig:env_last}. Here it can be seen that there are no significant differences between the three search algorithms.

\begin{figure*}
    \centering
    \includegraphics{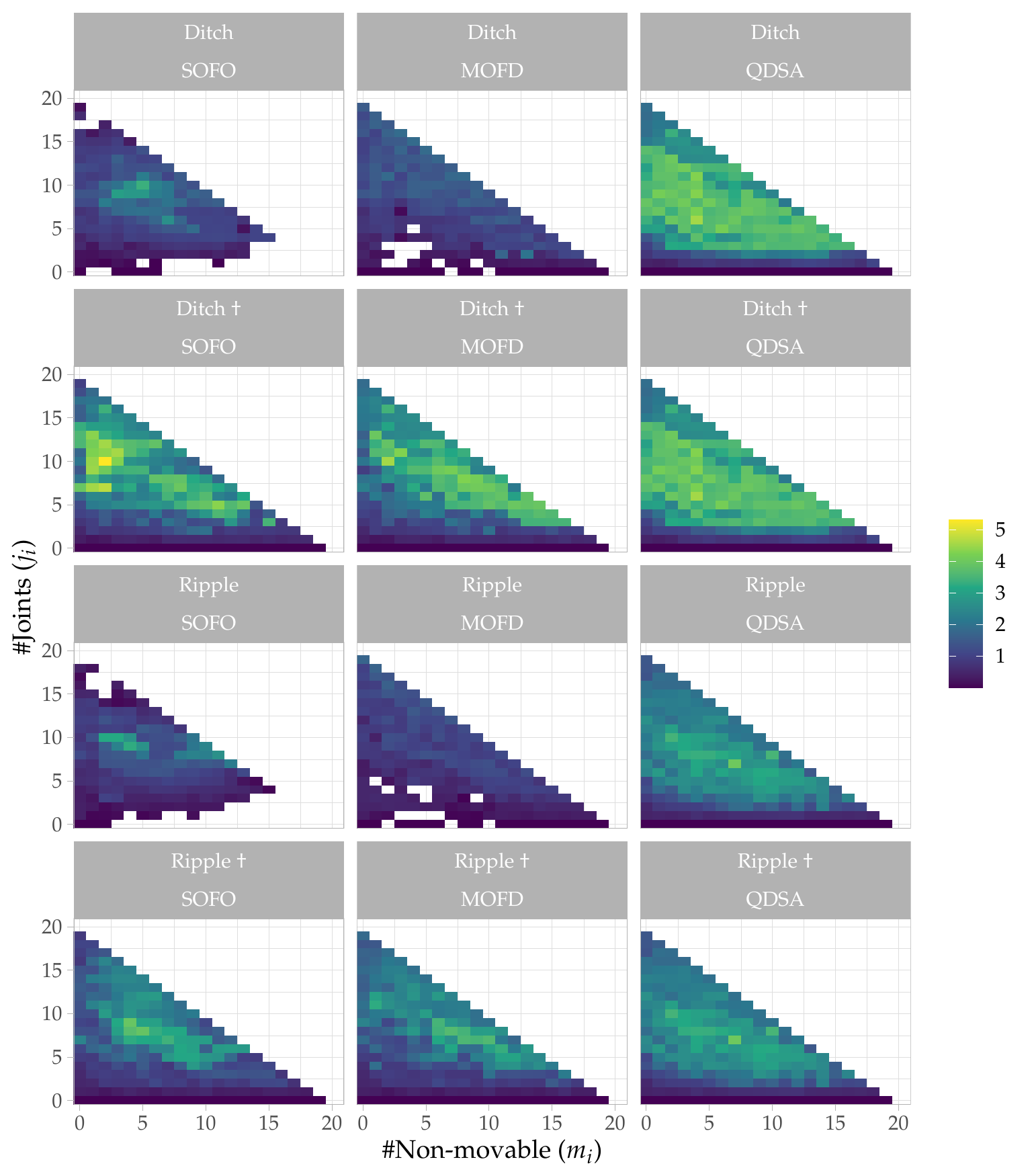}
    \caption{Population projection in the different environment settings. The projections shows the best solution for each morphological descriptor over all runs after the last evaluation. Environment names with $\dagger$ signify that both \gls{single} and \gls{multi} was initialized with the result of \gls{map} from the default flat environment.}
    \label{fig:env_maps}
\end{figure*}

To highlight the difference between the evolved populations in the different environmental settings, we projected the best found solutions for the different morphological descriptions in \cref{fig:env_maps}. This figure show that both \gls{single} and \gls{multi} are able to solve the more challenging environments when initialized with the result of \gls{map}. However, when started from their respective previous population from the flat environment they are not able to regain the same fitness as \gls{multi}. Note that the figures show the cumulative best solution which accounts for the large difference in number of filled cells for the $\dagger$ environments.

\subsection{Genealogical Analysis}
\begin{figure*}
    \centering
    \begin{subfigure}[b]{0.45\textwidth}
        \centering
        \includegraphics[height=2.5in]{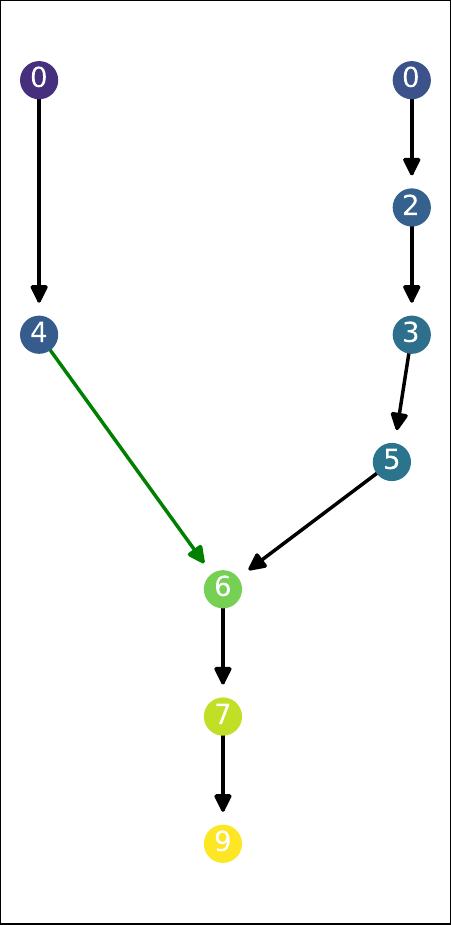}
    \end{subfigure}
    \begin{subfigure}[b]{0.45\textwidth}
        \begin{tikzpicture}
        \node[inner sep=0pt]{\includegraphics[height=2.5in]{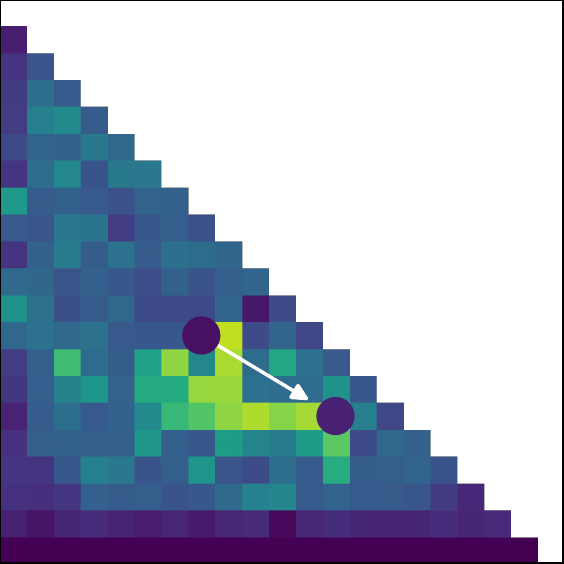}};
        \node[circle,draw=white, inner sep=0pt,minimum size=12pt, very thick] at (0.61, -1.51) {};
        \node[circle,draw=white, inner sep=0pt,minimum size=12pt, very thick] at (-0.91, -.61) {};
        \end{tikzpicture}
    \end{subfigure}
    \caption{Visualized ancestry of a solution. On the \textbf{left} the circles represents solutions that are ancestors of the bottommost circle. The color of the circle represents the fitness of the solution. The number is the generation when the solution appeared in the population. Arrows indicate parenthood, where black arrows indicate that there is \textit{no} morphological difference and green arrows indicate a morphological difference between parent and child. The two arrows joining to one new solution corresponds to a crossover operation while the other arrows correspond to mutation operations. On the \textbf{right} the tree is projected back into a repertoire and shows that the whole tree developed in just two different morphological descriptions, the white arrow indicates a change from parent to child in morphology. }
    \label{fig:example_ancestry}
\end{figure*}

Up until now the focus has been on the quality and diversity of the evolved populations, however, the previous graphs have not been able to show why the search algorithms evolve differently. We will therefore analyze the genealogical history of solutions to better understand how the solutions evolved. \Cref{fig:example_ancestry} shows an illustration of the genealogical ancestry that we will analyze. On the left the ancestry of a single solution is shown and on the right the same ancestry is projected into the repertoire it came from, showing the different morphological niches the ancestry occupies. The summary metrics which we will present is generated by taking each solution in final the population, extract the genealogical ancestry --- as shown in \cref{fig:example_ancestry}, apply a summary function to that ancestry and then collate the results of all the solutions.

\begin{figure*}
    \centering
    \begin{subfigure}[b]{\textwidth}
        \centering
        \includegraphics{cache/legend.pdf}
    \end{subfigure}
    \begin{subfigure}[b]{0.45\textwidth}
        \includegraphics{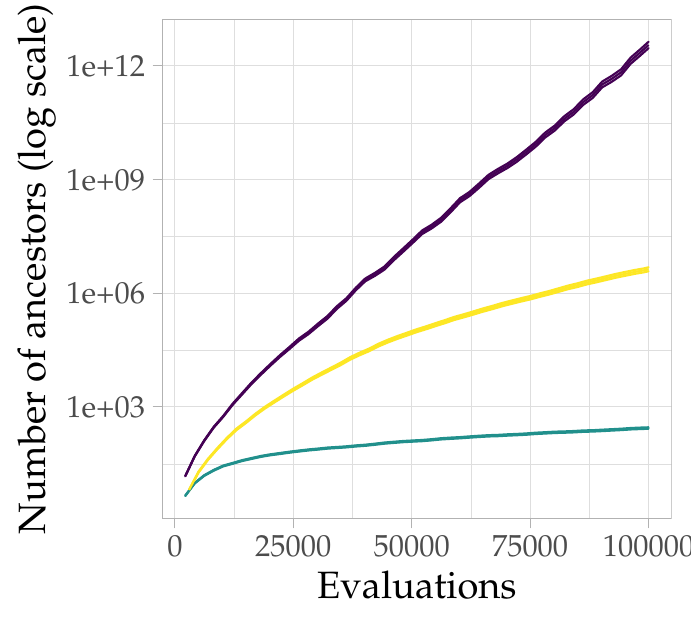}
    \end{subfigure}
    \begin{subfigure}[b]{0.45\textwidth}
        \includegraphics{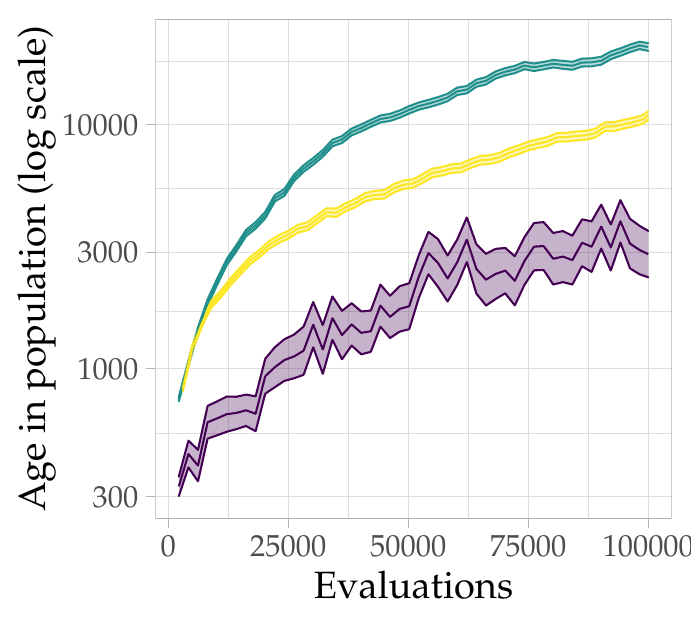}
    \end{subfigure}
    \caption{Genealogical ancestry attributes. On the \textbf{left} the average number of ancestors for each solution in the population is shown. On the \textbf{right} the age of each solution is shown, where age is number of evaluations since the solution first appeared in the population. The colored area represents the 95\% confidence interval.}
    \label{fig:genealogy1}
\end{figure*}

\Cref{fig:genealogy1} shows the number of ancestors and the age of solutions over generational time. 
The number of ancestors is simply the size of the ancestry, for the example in \cref{fig:example_ancestry} the number of ancestors would be $8$, and gives an indication of how often solutions are replaced. Another way to look at this replacement is to measure the age of solutions. In \cref{fig:genealogy1} age is measured as the number of evaluations since a solution appeared in the population. The two graphs illustrates that the generational replacement \gls{ea} creates a lot of new solutions, making the average age in the population very low, while the two other search algorithms tend to generate fewer new solutions and thus have a higher age.

\begin{figure*}
    \centering
    \begin{subfigure}[b]{\textwidth}
        \centering
        \includegraphics{cache/legend.pdf}
    \end{subfigure}
    \begin{subfigure}[b]{0.45\textwidth}
        \includegraphics{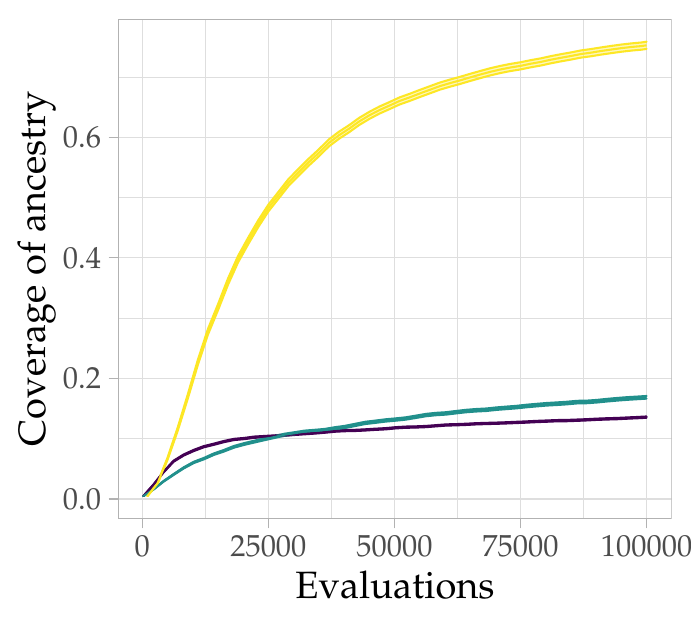}
    \end{subfigure}
    \begin{subfigure}[b]{0.45\textwidth}
        \includegraphics{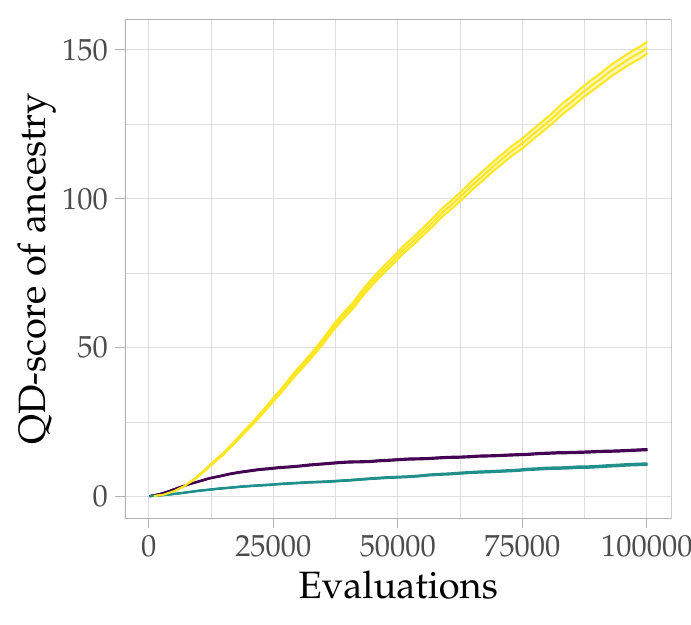}
    \end{subfigure}
    \caption{\acrlong{qd} metrics applied to the genealogical ancestry of solutions after the ancestry is projected into the repertoire as illustrated in \cref{fig:example_ancestry}. 
    On the \textbf{left}, coverage is shown, which is the number of morphological niches occupied by the ancestry. 
    On the \textbf{right}, the \gls{qd}-score is shown, which is a summation of the fitness of the ancestry after the ancestry has been projected into a repertoire.}
    \label{fig:genealogy_qd}
\end{figure*}

As shown in \cref{fig:example_ancestry}, projecting the ancestry into a repertoire of morphological niches can be a way to gain insight into how a solution evolved over time. In \cref{fig:genealogy_qd} the quality-diversity metrics, coverage and the QD-score, are applied to the ancestry of solutions in the population. 
This summary is different from the data in \cref{fig:qd_metrics} as this utilizes the genealogical ancestor tree and projects that into a repertoire, before applying the two metrics. 
The difference being that for the genealogical ancestry each solution in the final population is used to generate a unique repertoire consisting of only ancestors of the concluding solutions before \gls{qd} metrics are applied to each of these `ancestor repertoires'. 
\cref{fig:genealogy_qd} shows the coverage of ancestry (left), and on the \gls{qd}-score (right). 
From the figures it can be seen that the solutions in \gls{map} have an ancestry which covers a larger fraction of the morphological search space. 
This is contrasted with \gls{multi} which is able to obtain quite good coverage, as seen in \cref{fig:qd_metrics}, while the ancestry of solutions tends to have a much lower morphological diversity. 
One way to interpret this is that solutions in \gls{map} tend to share more ancestry with morphologically different solutions compared to \gls{multi}.

\begin{figure*}
    \centering
    \begin{subfigure}[b]{\textwidth}
        \centering
        \includegraphics{cache/legend.pdf}
    \end{subfigure}
    \begin{subfigure}[b]{\textwidth}
        \centering
        \includegraphics{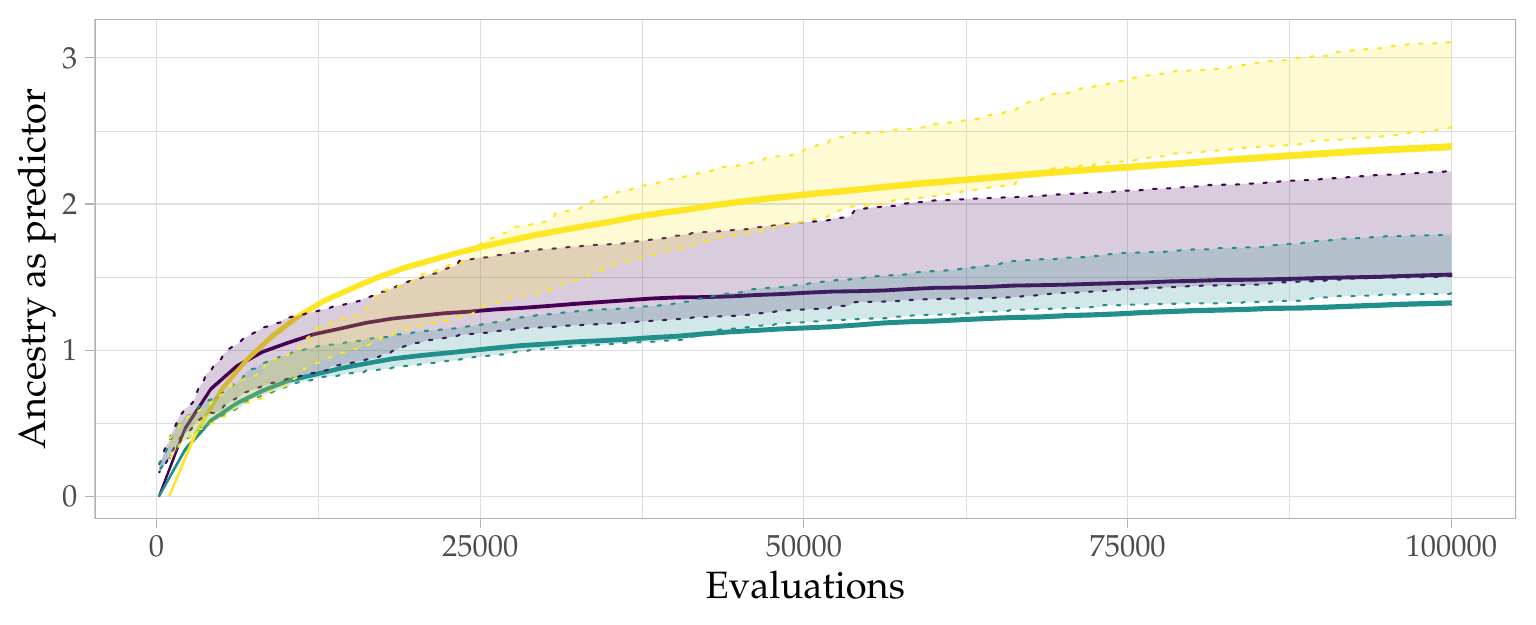}
    \end{subfigure}
    \caption{Using \acrlong{qd} metrics of ancestry to predict maximum fitness of each search algorithm. The dotted area represents the $95\%$ confidence interval of maximum fitness, as shown in \cref{fig:fitness}, and the line represents the estimated fitness based on a linear model of \acrlong{qd} metrics of ancestry. The linear model achieves an \textit{$R^2$} of $0.9084$.}
    \label{fig:genealogy_pred}
\end{figure*}

To test the prediction of whether or not the \gls{qd} metrics as applied to ancestry, shown in \cref{fig:genealogy_qd}, are good predictors of maximum fitness, we created a linear model. The linear model predicts maximum obtained fitness based on logarithmic coverage and \gls{qd}-score, both from ancestry. The model fit the data with an \textit{$R^2$} of $0.9084$, which indicates that the model fit the data quite well. To verify if the model fit with the maximum fitness of \cref{fig:fitness}, we plotted the $95\%$ confidence interval of the fitness, as shown in \cref{fig:fitness}, overlaid with the estimated fitness based on coverage and \gls{qd}-score in \cref{fig:genealogy_pred}. From the figure it can be seen that the model is challenged by the larger difference in ancestry between the three search algorithms, \cref{fig:genealogy_qd}, compared to the lower difference in maximum fitness, \cref{fig:fitness}. The figure illustrates that the model, for a large part of the data, match the obtained maximum fitness and thus could be an indication that these ancestry metrics are a good predictor of maximum fitness.

\section{Discussion}


The results for the locomotion task, in the default flat terrain, shows that \gls{map} produced the highest performing solutions for this problem (\cref{fig:fitness}). 
Our previous work~\citep{nordmoen2020quality} showed no quantifiable difference between the \gls{map}. 
For the experiments in this paper we added a curiosity score that led to increased performance of \gls{map} within our computational budget.
This corroborates the previous findings mentioned in \citep{cully2017quality} of curiosity being a useful addition to \gls{map-elites}. 
From the performance results, it can also be seen that \gls{map} produces the most diverse populations going as far as filling out all the morphological niches defined, \cref{fig:population_num}, and is the most consistent at finding high performing solutions, \cref{fig:population_avg}. 
\cref{fig:qd_metrics} shows that explicitly promoting diversity through either objective- or \gls{qd} based search algorithms improves the assortment of morphologies. 
However, \gls{map} produces more diversity compared to \gls{multi}, demonstrating the advantage of \gls{qd} algorithms for this task. 
Coupled together, it is likely that the diversity produced in \gls{map} together with the selection pressure to improve in promising areas of the search landscape lead to higher overall performance, underscoring the utility of diversity in \glspl{ea}. 
The effect of this could potentially explain the slower rise in performance seen on the left in \cref{fig:fitness} in contrast to the sharp growth of coverage on the left in \cref{fig:qd_metrics}. 
A potential explanation is that the search initially increases diversity, likely because it is easy to fill empty morphological niches, before being forced to improve performance of existing solutions. 
Improving performance can be done either by creating offspring with better controllers in the same position on the map or new morphologies that occupy a new position on the map. 
In both of these possibilities, the performance can only increase and a higher degree of diversity helps with creating new solutions that outperform existing solutions.

\cref{fig:env_last}  shows that \gls{map} is able to gain significantly higher performance when transitioning to new environments compared to the two other approaches.
However, all approaches achieved equal fitness when seeded with the population of \gls{map} at the time of transition.
The seeding also led to all three search algorithms finding good solutions in the same area of the search space (\cref{fig:env_maps}).
This shows the advantage of morphological diversity, as all three search algorithms are able to recover a large portion of fitness when initialized from a sufficiently diverse population. 
However, only the \gls{map} approach had managed to build up this level of diversity before the environmental change.
The reason for the similar performance when initialized from the \gls{map} population could be due to a highly deceptive fitness landscape.
This could mean that e.g. the \gls{ea} on its own would be struggling to find a suitable morphology, whereas the seeded population of diverse morphologies may already contain a body in a good location of the search space, leaving time for controller optimization. 
%
The results from the environment transitions highlight two important aspects of diversity.
Sufficient diversity is required to find solutions for difficult environments, and when experimenting with different environments in \gls{er}, it is difficult to a priori predict if a search algorithm is capable of finding a solution. 
However, search algorithms that produce and maintain diversity are more likely to handle the challenge.

The analysis of genealogical ancestry revealed several interesting aspects about how the population of solutions evolve for the three different search algorithms. Based on the number of ancestors and age in population, \cref{fig:genealogy1}, it can be seen that the generational replacement aspect of \gls{single} lead to many young solutions compared to the two other search algorithms. This is to be expected and does not appear to disadvantage \gls{single} compared to \gls{multi} in regards to performance. When looking at the difference between \gls{multi} and \gls{map} it can be seen that \gls{map} is producing more solutions throughout evolution. One potential explanation for this is that due to the complex Pareto dominance calculated for solutions in \gls{multi} solutions are rarely replaced based on fitness and once diversity is maximized, as illustrated on the left in \cref{fig:qd_metrics}, the search stagnates. This shows the complexities of introducing diversity into a maximization regime and it is likely that the two additional objectives are reducing the opportunity to improve on fitness. Based on the \gls{qd} metrics of the ancestry, shown in \cref{fig:genealogy_qd}, it can be seen that ancestors of solutions in \gls{map} cover a large area of the search space compared to the two other search algorithms. One could expect that since \gls{single} has many orders of magnitude more ancestors the solutions would cover a large area of the search space, however, the \gls{qd} metrics show that the ancestors are not as morphological diverse as in \gls{map}. This result builds on the notion that \gls{map-elites} is better at generating diverse stepping stones, as proposed by \citet{mouret2015illuminating}. This is also underscored by the \gls{qd}-score, on the right in \cref{fig:genealogy_qd}, which show that\textemdash in general, ancestors in \gls{map} are both diverse and high performing. Lastly, by building a linear model, predicting maximum fitness based only on coverage and \gls{qd}-score of ancestry, we showed that the notion of stepping stones could be a good predictor of performance. By modelling the relationship between genealogical ancestry and search performance, stepping stones can be seen in the larger context of evolution and gives an even stronger indication that \gls{map-elites} is able to produce impressive results based on diverse and high performing intermediary solutions. By performing this analysis across populations, over many runs, we are able to gain statistical insight into the notion of stepping stones which strengthens the overall conclusion.


We showed how \gls{map} is an effective method for evolving both the morphology and control of modular robots for performance, diversification, and transfer to new environments. 
While the implementation of \gls{map} is promising for evolving both the morphology and control of modular robots, there are additional challenges related to how different selection methods could be further improved in \gls{map}.
The modular robotics approach furthermore allows us to expand our module inventory to incorporate various types of other structural, sensing and actuator modules. 
This possible extension will convolute the search space further and possibly benefits more from \gls{map} than other algorithms. 
A challenge is that different genotypes can still map to the same morphological features, which might be a problem as it can constrain the type of solutions found to particular robot morpholgies.
Therefore, additional morphological features could be implemented to create a multi-dimensional map that could lead to better and more unique solutions. 
Since adding morphological features also increases computational demand (there are more cells to explore), \gls{map} might be combined with other extensions such as Covarianve Matrix Adaptation MAP-Elites \citep{Fontaine_2020}.

\section{Conclusion}
Optimizing both the morphology and controller for evolving modular robots is challenging due to the large and unknown search space.
As the amount of exploration vs exploitation to be used for optimization strategies is usually determined by the complexity of the agent and the environment\textemdash the ruggedness of the resulting fitness landscape\textemdash we compared three evolutionary algorithms to determine how each performs on this challenging search space.
From the results of the three implemented evolutionary algorithms the \acrlong{map} algorithm produced comparably higher performing and more morphologically diverse solutions compared to \acrlong{single} and \acrlong{multi}. 
The importance of the diversity produced by \acrlong{map} is corroborated through experiments on transferring evolved robots to two additional, and more difficult, terrain types.
The genealogical ancestry produced by each algorithm furthermore indicated that \acrlong{map} found more stepping stones, shedding new light on how the algorithm achieves high-performing final solutions.
The added pressure for diversification in the two morphological dimensions of the implemented \acrlong{map} bolsters how useful it can be for evolving both the morphology and control of robots.

\section*{Author Contributions}
Idea and experiment design developed by \textit{Jørgen Nordmoen}, \textit{Frank Veenstra}, \textit{Kai Olav Ellefsen} and \textit{Kyrre Glette}. Robot simulation framework developed by \textit{Jørgen Nordmoen} and \textit{Frank Veenstra}. Experimental code and experiments run by \textit{Jørgen Nordmoen}. All the authors contributed to the analysis of the experiments and the writing of the article.

\section*{Funding}
This work is partially supported by The Research Council of Norway through its Centers of Excellence scheme, project number 262762

\section*{Acknowledgments}
The simulations were performed on resources provided by UNINETT Sigma2 - the National Infrastructure for High Performance Computing and Data Storage in Norway.

\section*{Supplemental Data}
 Supplementary material can be found at \url{\homepage}.

\printglossary[type=\acronymtype,title=Abbreviations]

\bibliographystyle{frontiersinSCNS_ENG_HUMS} 
\bibliography{modular}

\end{document}

%% file: glossary.tex
\newacronym[firstplural=Evolutionary Strategies (ES)]{es}{ES}{Evolutionary Strategy}
\newacronym{ann}{ANN}{Artificial Neural Network}
\newacronym{cma-es}{CMA-ES}{Covariance Matrix Adaptation - Evolutionary Strategy}
\newacronym{cpg}{CPG}{Central Pattern Generator}
\newacronym{dyret}{DyRET}{Dynamic Robot for Embodied Testing}
\newacronym{ea}{EA}{Evolutionary Algorithm}
\newacronym{ec}{EC}{Evolutionary Computation}
\newacronym{er}{ER}{Evolutionary Robotics}
\newacronym{grf}{GRF}{Ground Reaction Force}
\newacronym{hpc}{HPC}{High Performance Computing}
\newacronym{map-elites}{MAP-Elites}{Multi-dimensional Archive of Phenotypic Elites}
\newacronym{moea}{MOEA}{Multi-Objective Evolutionary Algorithm}
\newacronym{nsga}{NSGA-II}{Non-dominated Sorting Genetic Algorithm-II}
\newacronym{nslc}{NSLC}{Novelty Search with Local Competition}
\newacronym{ns}{NS}{Novelty Search}
\newacronym{qd}{QD}{Quality Diversity}
\newacronym{rem}{REM}{Robotics, Evolution and Modularity}
\newacronym{ros}{ROS}{Robot Operating System}
\newacronym{supg}{SUPG}{Single-Unit Pattern Generator}

\newacronym{single}{SOFO}{Single Objective Fitness Only}
\newacronym{multi}{MOFD}{Multi Objective Fitness \& Diversity}
\newacronym{map}{QDSA}{Quality Diversity with Structured Archive}

%% file: figures/encoding.tex
\begin{tikzpicture}
\node[rectangle, label=above:Root, minimum size=1.5cm, draw=blue](root1) {Rect};
\node[rectangle, minimum size=1.5cm, draw=teal, below left=1cm of root1](serv1) {Servo}
    edge[<-, edge label={$(1, 0, 0)$}](root1.south);
\node[rectangle, minimum size=1.5cm, draw=teal, below right=1cm of root1](serv2) {Servo}
    edge[<-, edge label={$(0, 1, 0)$}, swap](root1.south);
\node[rectangle, minimum size=1.5cm, draw=blue, below=1cm of serv2] {Rect}
    edge[<-, edge label={$(0, 0, 1)$}](serv2.south);
\end{tikzpicture}